\documentclass[]{bytedance_seed}

\usepackage[toc,page,header]{appendix}

\usepackage{wrapfig}
\usepackage[most]{tcolorbox}

\newtcolorbox{underlinetextbox}{
    boxrule=0pt, frame hidden, colback=white,
    borderline south={1pt}{0pt}{black} 
}

\usepackage{minitoc}

\usepackage{times}
\usepackage{latexsym}

\usepackage[T1]{fontenc}

\usepackage[utf8]{inputenc}

\usepackage{microtype}

\usepackage{inconsolata}

\usepackage{graphicx}
\usepackage{booktabs}
\usepackage[table]{xcolor} 
\usepackage{tabularx}      
\usepackage{ragged2e} 
\usepackage{multirow}
\usepackage{tcolorbox}
\usepackage{verbatim}
\usepackage{enumitem}
\usepackage{amsmath} 
\usepackage{tipa}
\usepackage{threeparttable}

\title{CoTJudger: A Graph-Driven Framework for Automatic Evaluation of Chain-of-Thought Efficiency and Redundancy in LRMs}

\author[1,4 *]{Siyi Li}
\author[2,8 *]{Jiajun Shi}
\author[3\dagger]{Shiwen Ni}
\author[2,\dagger]{Ge Zhang}
\author[4]{Shuaimin Li}
\author[5]{Shijian Wang}
\author[2]{Zhoufutu Wen}
\author[9]{Yizhi LI}
\author[6]{Hamid Alinejad-Rokny}
\author[7,\dagger]{Jiaheng Liu}
\author[3,4\dagger]{Min Yang}
\author[2]{Wenhao Huang}

\affiliation[1]{University of Science and Technology of China}
\affiliation[2]{ByteDance SEED}
\affiliation[3]{Shenzhen University of Advanced Technology}
\affiliation[4]{Shenzhen Institutes of Advanced Technology, CAS}
\affiliation[5]{Southeast University}
\affiliation[6]{UNSW Sydney}
\affiliation[7]{Nanjing University}
\affiliation[8]{Beihang University}
\affiliation[9]{University of Manchester}

\contribution[*]{Equally contributed authors}
\contribution[\dagger]{Corresponding authors}

\abstract{
Large Reasoning Models (LRMs) have demonstrated strong performance by producing extended Chain-of-Thought (CoT) traces before answering. However, this paradigm often induces \emph{over-reasoning}: redundant calculations and circular self-verification that increase computational cost without improving outcomes. Existing evaluations largely emphasize final accuracy or coarse token counts, and lack automated tools to separate essential logic from structural redundancy. We introduce \textbf{CoTJudger}, a graph-driven framework that quantifies reasoning efficiency by converting free-form CoTs into directed dependency graphs and extracting the \emph{Shortest Effective Path} (SEP) needed to reach a correct solution. This yields an interpretable efficiency signal---how much of a CoT is necessary versus structurally redundant---that is comparable across models and tasks. Evaluating 21 LRMs, CoTJudger reveals pervasive redundancy and surfaces recurring failure modes, including \emph{verification obsession} and \emph{compensatory redundancy}. These results provide a practical metric for disentangling reasoning ability from computational waste, enabling more targeted evaluation and diagnosis of LRM efficiency.

}

\date{\today}
\correspondence{Siyi Li at \email{sa24225212@mail.ustc.edu.cn}, Jiajun Shi at \email{shijiajun.123@bytedance.com}, Shiwen Ni at \email{nishiwen@suat-sz.edu.cn}, Ge Zhang at \email{zhangge.eli@bytedance.com}}

\checkdata[Project Page]{\url{https://github.com/41ForOne/CoTJudger}}

\begin{document}
\maketitle


\section{Introduction}

\begin{figure}[t]
  \includegraphics[width=\linewidth]{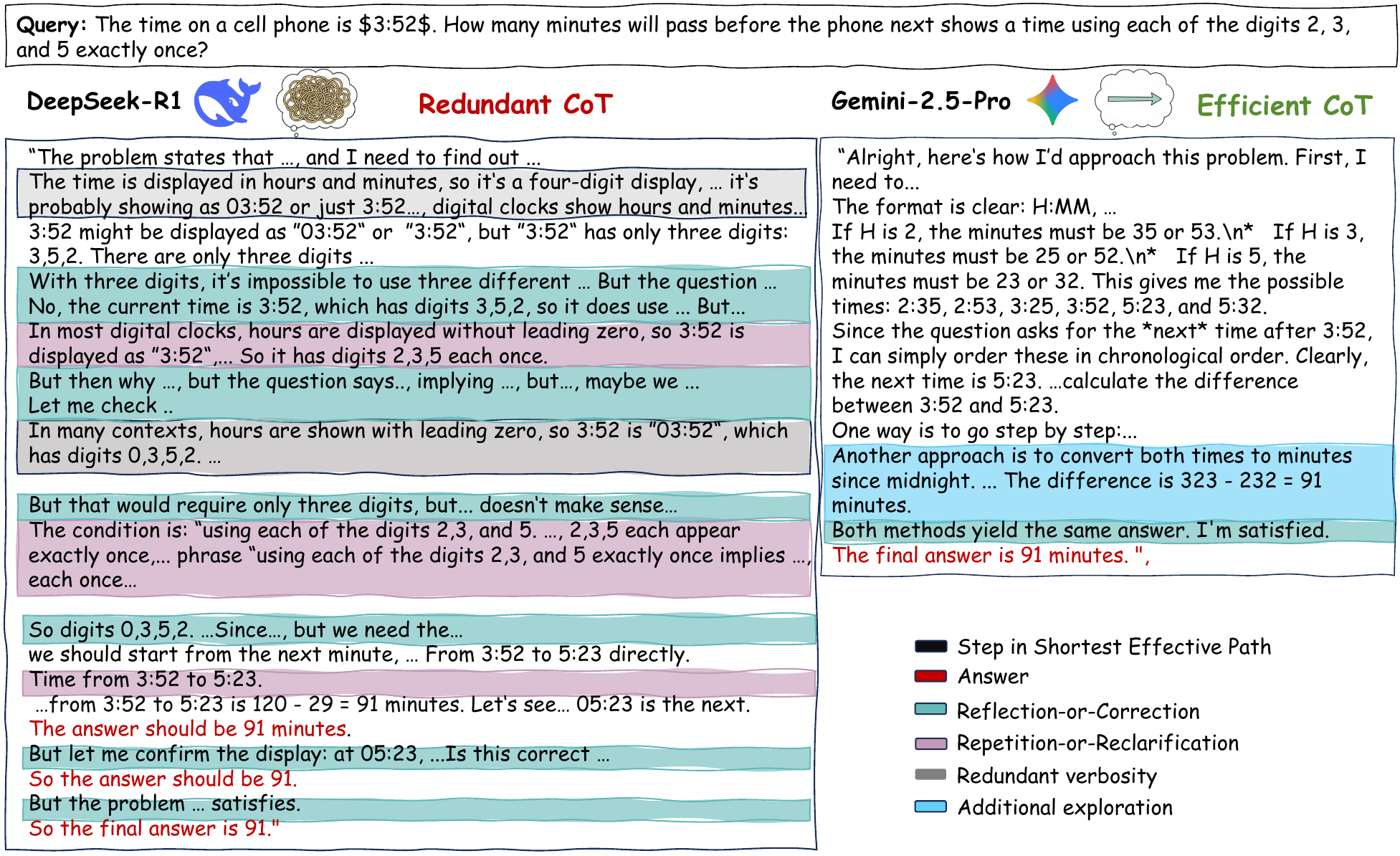}
\caption{Two Chain-of-Thought (CoT) traces from DeepSeek-R1 and Gemini-2.5-Pro on a temporal reasoning task. Although both models reach the correct answer, DeepSeek-R1 (left) shows substantial verbosity, repetition, and multiple reflection/correction loops. In contrast, Gemini-2.5-Pro (right) follows a more direct and efficient path with minimal additional exploration.}

  \label{fig:examples}
\end{figure}

Large Reasoning Models (LRMs) such as OpenAI o1 \citep{jaech2024openai}, DeepSeek-R1 \citep{guo2025deepseek}, and Kimi-K2-Thinking \citep{kimiteam2025kimik2} represent a major shift in how large language models solve complex tasks. While long-form Chain-of-Thought (CoT) can improve performance, it also introduces substantial inference overhead. This creates a practical tension: extended reasoning often drifts into over-reasoning, with superfluous computations, unproductive backtracking, and circular self-verification (Figure~\ref{fig:examples}) \citep{chiang-lee-2024-reasoning, sui2025stop, chen2024not}.

Current approaches to evaluating CoT redundancy primarily rely on token-based signals. Methods such as OptimalThinkingBench \citep{aggarwal2025optimalthinkingbenchevaluatingunderthinkingllms}, O1-Pruner's AES \citep{luo2025o1}, and CoT-Valve's ACU \citep{ma-etal-2025-cot} relate CoT length to accuracy and model size, while benchmarks such as DNABench \citep{hashemi2025dnrbenchbenchmarkingoverreasoning} target excessive generation under adversarial settings. However, these coarse metrics cannot distinguish necessary complexity from structural waste. In contrast, PRMBench \citep{song2025prmbenchfinegrainedchallengingbenchmark} provides step-wise annotations, but the process is costly and difficult to scale. As a result, the field lacks an automated structural framework that separates circuitous verbosity from concise efficiency, creating a risk of optimizing for token volume rather than reasoning quality.

We address this gap with a simple viewpoint: \textit{reasoning efficiency can be measured by how far a CoT deviates from the shortest logically coherent path from the problem to the answer}. Such deviations can appear as verbosity, misguided exploration, excessive self-verification, or inefficient correction loops. We operationalize this viewpoint with \textbf{CoTJudger}, a graph-driven framework for automated redundancy evaluation. Because linear text obscures branching, looping, and self-correction, we convert unstructured CoTs into directed dependency graphs, where nodes represent atomic functional steps and edges encode logical dependencies. This representation turns verbosity into computable topological structure, enabling us to extract the Shortest Effective Path (SEP) and separate core reasoning from redundant segments.

We evaluate 21 LRMs across math, programming, PCB (physics, chemistry, biology), and general reasoning tasks, and find redundancy to be widespread. Beyond aggregate scores, our multidimensional attribution analysis links redundancy to step position, task difficulty, and error patterns, and highlights recurring failure modes that inflate compute without improving outcomes. The contributions of this paper are fourfold:
\begin{itemize}
\item We introduce \textbf{CoTJudger}, a \emph{structure-aware} evaluator that converts free-form CoTs into directed dependency graphs and algorithmically extracts the \textit{Shortest Effective Path} as the minimal reasoning skeleton leading to a correct answer.
\item We propose a domain-agnostic \textbf{Functional Node Classification System} that maps CoT spans to atomic step types (e.g., \textit{Problem-Deconstruction}, \textit{Reflection-or-Verification}), enabling interpretable attribution of redundancy to specific reasoning behaviors.
\item We perform a large-scale study of 21 LRMs and identify recurring redundancy patterns, including \textit{Verification Obsession}, \textit{Compensatory Redundancy}, and \textit{Logical Epicenters}, providing empirical evidence of how over-reasoning manifests across tasks.
\item We define the \textbf{Redundancy Ratio ($R$)} as a structural efficiency metric grounded in the SEP, offering a scalable objective for comparing LRMs and guiding efficient-reasoning methods such as reward modeling.
\end{itemize}

\section{Related Work}

\noindent \textbf{Efficiency and Redundancy in Large Reasoning Models.}
Motivated by the growing inference cost of LRMs and the ``overthinking'' behavior seen in long CoTs, recent work has targeted redundancy through minimalist-token strategies such as Chain-of-Draft \citep{xu2025chaindraftthinkingfaster} and training- or pruning-based methods \citep{zhao2025letlrmsbreakfree, luo2025o1} \citep{chiang-lee-2024-reasoning, feng2025efficient}. However, these approaches largely optimize length or stopping criteria rather than diagnosing the structural sources of redundancy \citep{liu2025learn, cheng2025optimizing}. CoTJudger bridges this gap via a diagnostic framework designed to quantify both the source and drivers of inefficiencies.

\noindent \textbf{Graph-Based Automatic Reasoning Evaluation.}
Recent white-box approaches that analyze the reasoning process, such as Circuit-based Reasoning Verification (CRV) \citep{zhao2025verifyingchainofthoughtreasoningcomputational}, identify errors by leveraging internal attribution graphs. Flow of Reasoning (FoR) \citep{yu2025flowreasoningtrainingllms} similarly models reasoning as a Markovian flow on a Directed Acyclic Graph (DAG). CoTJudger complements these lines of work by bringing a graph-based view to external, text-based CoT evaluation: by mapping textual logic into a computable dependency graph, it enables automated and interpretable assessment of reasoning paths at a scale that is difficult to achieve with human-annotated benchmarks such as MME-CoT \citep{jiang2025mmecotbenchmarkingchainofthoughtlarge}.
\section{Data Description}
\noindent \textbf{Query Dataset.}
We constructed a balanced dataset of 896 queries from open-source benchmarks spanning four domains: Math (364), General Reasoning (270), Programming (164), and PCB (98). To promote coverage, we applied stratified sampling using subdomain labels and difficulty annotations from the source datasets. Additional construction details are provided in Appendix~\ref{sec:query_details}.

\noindent \textbf{Core Development Set.}
To improve robustness and reduce reliance on ad-hoc heuristics, we developed our framework (Section~\ref{sec:cotjudger}) through a data-driven process. We curated a diverse core development set of 2,688 CoTs generated by three representative models: Gemini-3-Pro (proprietary SOTA), Qwen3-30B-A3B-Thinking (open-source LRM), and DeepSeek-R1-Llama-70B (distilled). This set supported iterative module-level validation and tuning, grounding the framework in large-scale empirical evidence rather than subjective design choices and improving generalization to unseen models. Detailed analyses and experiments are provided in Appendix~\ref{sec:development}.

\section{CoTJudger}
\label{sec:cotjudger}
\subsection{Step Segmentation and Atomization}
\label{subsec:segmentation}

\begin{figure*}[t]
  \includegraphics[width=\linewidth]{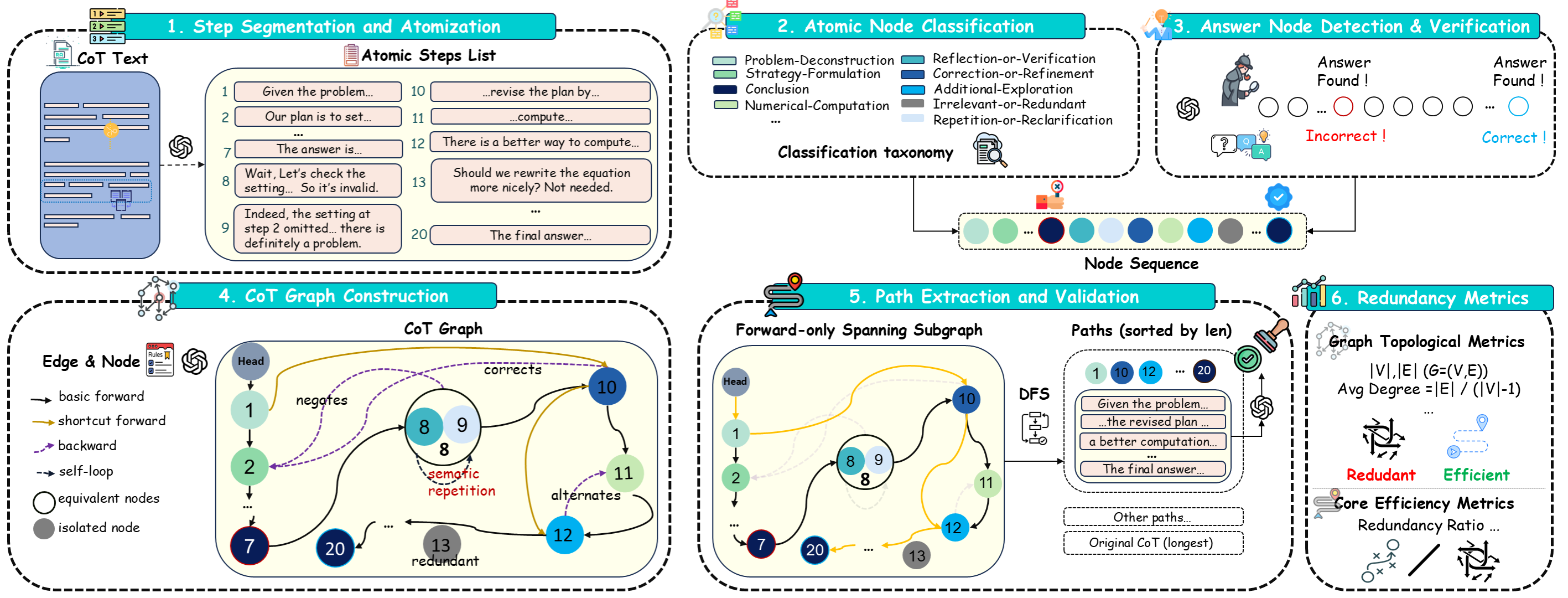} 
  \caption{Automatic evaluation framework of \textbf{CoTJudger}. The pipeline comprises six modules: (1) Step Segmentation and Atomization, (2) Atomic Node Classification, (3) Answer Node Detection and Verification, (4) CoT Graph Construction, (5) Path Extraction and Validation, and (6) Redundancy Metrics Calculation.}

  \label{fig:overview}
\end{figure*}

To transform free-form CoTs into atomic logical units, we use a two-stage pipeline.

\noindent \textbf{Initial Heuristic Segmentation.}
We first perform coarse-grained segmentation by detecting the most frequent delimiter (line breaks, \texttt{\textbackslash n}, or double line breaks, \texttt{\textbackslash n\textbackslash n}). To preserve syntactic integrity, we apply protective masking to code blocks before splitting. This yields an initial sequence $S = [s_1, s_2, \dots, s_k]$.

\noindent \textbf{LLM-based Atomization.}
To standardize granularity, we use GPT-5 to merge over-fragmented steps and split nodes that contain multiple reasoning actions. To avoid paraphrasing noise, the model outputs index-level structural edits rather than rewriting the text. This produces the final node list $\mathcal{V} = \{N_1, N_2, \dots, N_n\}$.

\subsection{Atomic Node Classification}
\label{subsec:classification_verification}

To determine the functional role of each node within the CoT structure, we adopt a unified two-tier taxonomy.

\noindent \textbf{Universal Two-Tier Classification System.}
To avoid dataset-specific and ad-hoc definitions, we construct a unified two-tier taxonomy that covers Math, General Reasoning, PCB, and Programming. We develop it via a systematic, data-driven pipeline on our core development set, combining bottom-up induction from domain-specific samples with top-down cross-domain aggregation. We then iteratively refine the taxonomy to disambiguate closely related concepts, such as separating evaluative \textit{Reflection-or-Verification} from executive \textit{Correction-or-Refinement}. The resulting taxonomy provides a robust basis for automated analysis of CoT structure. See Appendix~\ref{sec:appendix_taxonomy} for the full taxonomy and construction details.

\noindent \textbf{Automated Node Classification.}
Given the nodes $\mathcal{V}$ from Section~\ref{subsec:segmentation}, we use GPT-5 to assign labels under our two-tier taxonomy. The model is instructed to infer each node's contextual function within the global reasoning flow, reducing errors when surface phrasing (e.g., verification-like language) does not match the structural role (e.g., \textit{Intermediate-Inference}). This module outputs labeled nodes $\mathcal{V} = \{(N_i, l_i)\}$, which form the vertex set for subsequent CoT graph construction.

\subsection{Answer Node Detection and Verification}
Observations from the core development set show that some CoTs state an answer before the final node or present multiple candidate answers (correct or incorrect). To capture this information for evaluating redundancy and validity, we use GPT-5 to detect nodes that contain conclusive answers. For each candidate, we extract the proposed solution and verify it with domain-specific protocols (e.g., executing generated code in an isolated environment for programming tasks).

\subsection{CoT Graph Construction}
\label{subsec:graph_construction}

Linear text sequences cannot represent complex logical and meta-cognitive behaviors such as backtracking, repetition, or off-topic steps. To capture these dynamics, we model each CoT as a directed graph $G=(V,E)$, drawing inspiration from Control Flow Graphs (CFGs) and redundancy-aware structures in classic data representations. This formulation explicitly encodes logical dependencies and enables compression of redundant reasoning trajectories.

\noindent \textbf{Node Normalization.}
Before edge construction, we normalize the atomic nodes $\mathcal{V}$. For CoTs containing nodes classified as \textit{Repetition-or-Reclarification}, we use GPT-5 to detect semantic equivalence; nodes with high semantic overlap are assigned the same identifier. We also prepend a \textbf{Virtual Head Node} ($N_{\text{root}}$) as a unified origin to avoid boundary conditions during reconnection.

\noindent \textbf{Edge Construction Rules.}
We define three edge types (forward, backward, and self-loop) guided by node categories. A \textbf{Basic Forward} edge is the sequential link $(N_i, N_{i+1})$ added for all non-isolated nodes (i.e., nodes not labeled as \textit{Irrelevant-or-Redundant}). A \textbf{Self-loop} is added for duplicate nodes to denote semantic repetition. To represent meta-cognition, we add category-specific \textbf{Backward} and \textbf{Shortcut Forward} edges:
(1) \textbf{\textit{Correction-or-Refinement}}: If $N_s$ corrects $N_t$, we add a backward edge $(N_s, N_t)$ and a shortcut forward edge $(N_t.\mathrm{prev}, N_s)$ to bypass the erroneous subpath. (2) \textbf{\textit{Verification-or-Reflection}}: If $N_s$ verifies or reflects on $N_t$, we add $(N_s, N_t)$. If $N_s$ affirms $N_t$, a shortcut $(N_s.\mathrm{prev}, N_s.\mathrm{next})$ skips the verification node as auxiliary. If $N_s$ negates $N_t$, we add a shortcut $(N_t.\mathrm{prev}, N_s.\mathrm{next})$ to jump over the invalidated reasoning. (3) \textbf{\textit{Additional-Exploration}}: For alternative strategies or exploratory attempts, a backward edge $(N_s, N_t)$ links the new branch to the prior approach, with a shortcut $(N_t.\mathrm{prev}, N_s)$ denoting a logical jump to the alternative.

\subsection{Path Extraction and Validation}
\label{subsec:path_extraction}

Given $G$, multiple paths may exist between $N_{\text{root}}$ and each verified correct answer node $N_{\text{ans}}$. We define the \textbf{Shortest Effective Path (SEP)} as the shortest, logically self-consistent reasoning sequence that suffices to reach the answer.

To extract candidate paths, we first derive a spanning subgraph $G_{\text{forward}} \subset G$ by retaining only forward edges (basic and shortcut), which preserves paths that advance the solution. We then enumerate all paths from $N_{\text{root}}$ to each $N_{\text{ans}}$ in $G_{\text{forward}}$ using Depth-First Search (DFS) and sort them by node length. Finally, we concatenate the text of nodes in each path and use GPT-5 to verify whether the final answer can be rigorously derived from those steps alone. The first path that passes this validation is designated as the SEP.

For illustration, we provide two representative examples processed by the evaluation pipeline in Appendix~\ref{sec:cot_examples}. Prompts for each module are in Appendix~\ref{sec:prompt}.

\subsection{Redundancy Metrics Calculation}
\label{subsec:metrics}

\noindent \textbf{Basic Statistical Metrics.}
We report the total number of tokens in the CoT ($tokens$) and the accuracy rate ($acc$).

\begin{wrapfigure}{r}{0.5\columnwidth}
  \centering
  \includegraphics[width=0.5\columnwidth, keepaspectratio]{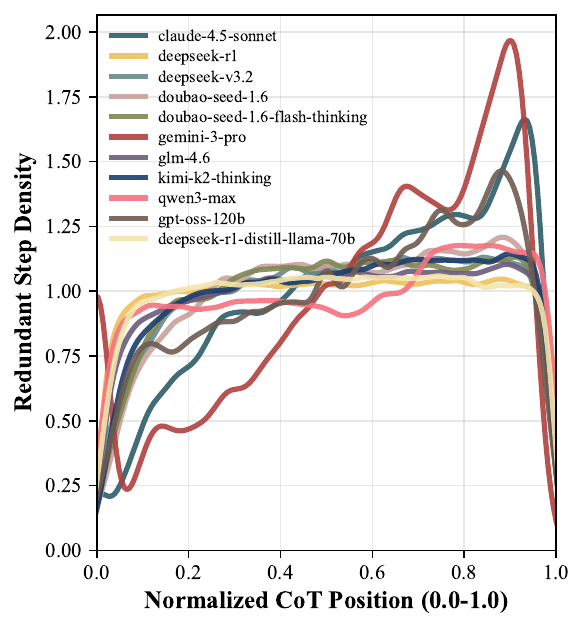}
  \caption{Positional distribution of redundant reasoning steps in CoT (KDE). The plot shows the normalized probability density of steps outside the Shortest Effective Path across models.}
  \label{fig:redundant_position_kde}
\end{wrapfigure}

\noindent \textbf{Graph Topological Metrics.}
These metrics quantify the structural density of a CoT. Beyond the basic counts of nodes $|V|$ and edges $|E|$, we define the \textbf{Isolated Node Ratio} as the fraction of nodes labeled \textit{Irrelevant-or-Redundant} that can be removed without affecting logical integrity, capturing content looseness and verbosity. We also compute \textbf{Edge Ratios} (the distribution of edge types) to measure the prevalence of advancement, backtracking, and repetition.
We define the \textbf{Average Degree} as $\overline{D}=\frac{|E|}{|V|-1}$. The amount by which $\overline{D}$ exceeds 1 quantifies topological overhead. Values near 1.0 correspond to an ideal CoT that follows the SEP without triggering reconnection rules $(|E|=|V|-1)$, whereas larger values indicate denser dependency structure. Unlike traditional graph density, $\overline{D}$ supports scale-invariant comparisons across CoTs of different lengths and directly reflects the average logical burden carried by a node.
We further identify \textbf{Logical Epicenters} using the \textbf{maximum in-degree and out-degree} $(indegree_{\max}, outdegree_{\max})$ to capture extreme local redundancy. High values indicate concentrated failure points where substantial path rewriting was needed to bypass erroneous or unnecessary segments.

\noindent \textbf{Core Efficiency Metrics.}
To quantify how much of a reasoning chain is functionally necessary for its final solution, we define: (1) \textbf{Shortest Effective Path Length} ($L_{\text{eff}}$), the number of nodes in the SEP; (2) \textbf{Redundancy Ratio} $\big(R=\frac{|V|-L_{\text{eff}}}{|V|}\big)$, the proportion of non-essential steps in the CoT; and (3) \textbf{Uncertainty Ratio} ($U$), the proportion of CoTs that contain more than two candidate answer nodes, serving as a proxy for decision inefficiency and wandering during reasoning.

Together, these metrics enable quantitative comparisons of efficiency across models and help localize redundancy patterns, providing interpretable signals for improving reasoning behavior.


\section{Experiment}
\subsection{Experimental Setup}

\begin{figure}[t]
\centering
  \includegraphics[width=0.8\columnwidth]{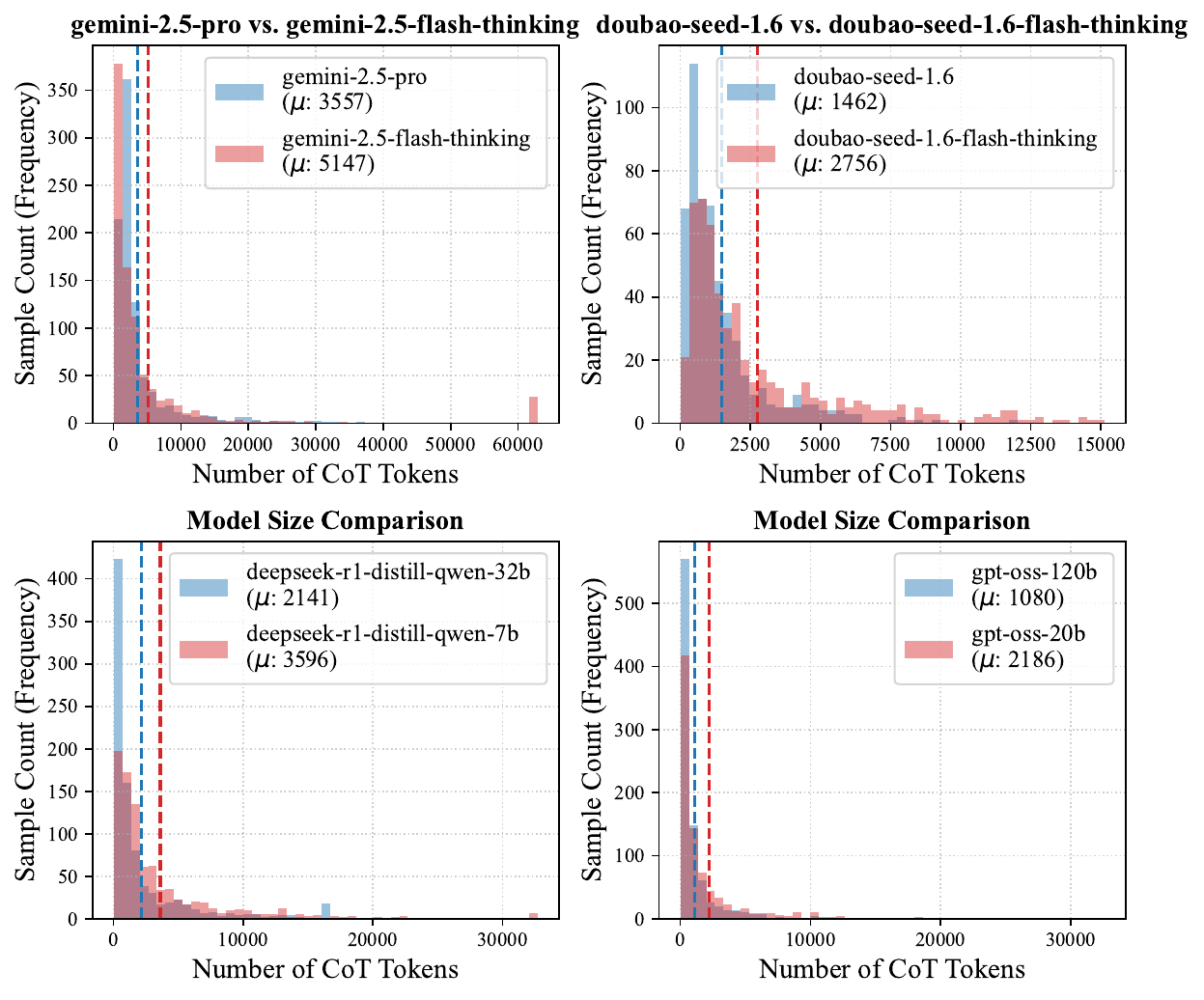}
  \caption{Comparative analysis of CoT token-length distributions, examining the effects of model variants (Pro/Base vs.\ Flash-Thinking) and parameter scaling on reasoning redundancy.}
  \label{fig:token_distribution_comparison}
\end{figure}
To evaluate LRM redundancy under different paradigms, we tested 21 models in our framework across three categories: \textbf{Proprietary LRMs}, including Claude-Sonnet-4.5 \citep{anthropic2025claude45}, the Gemini series \citep{comanici2025gemini}, the Doubao series \citep{doubao2025seed16}, and Qwen3-Max \citep{yang2025qwen3}; \textbf{Open-Source LRMs}, including GLM-4.6 \citep{zhipuai2025glm46}, the DeepSeek series \citep{guo2025deepseek,liu2025deepseek}, the Qwen series \citep{yang2025qwen3}, gpt-oss (20b, 120b) \citep{openai2025gptoss120bgptoss20bmodel}, and Kimi-K2-Thinking \citep{kimiteam2025kimik2}; and \textbf{Distilled LRMs} \citep{guo2025deepseek}, including multiple scale variants distilled from DeepSeek-R1.

We used greedy decoding without system instructions for all models to standardize generation, adapting CoT extraction (e.g., API fields versus prompt triggers) to each model interface (details in Appendix~\ref{sec:setup}). After answer verification, we re-stratified queries into five difficulty levels based on model performance to examine how difficulty correlated with CoT redundancy.

\subsection{Main Evaluation Results}
\label{subsec:overall_performance}
Table~\ref{tab:benchmark_results} reported results for 21 representative models (each corresponding to 896 CoT samples). By synthesizing the metrics in Section~\ref{subsec:metrics}, we presented a holistic assessment of CoT redundancy across LRMs.

\paragraph{The Efficiency Gap: Divergence Between Budget and Necessity.}
We observed a pronounced mismatch between how LRMs allocated computation and the complexity required by the query. DeepSeek-R1-0528-Qwen3-8B variants followed an aggressive expansion strategy, averaging 8,817 tokens (303 steps) per query. This contrasted sharply with the succinctness of Claude-Sonnet-4.5 (451 tokens) and Gemini-3-Pro (11 steps). Crucially, the Shortest Effective Path length ($L_{\text{eff}}$) indicated that the core reasoning needed for these queries averaged only 7--47 steps. As a result, models such as Qwen3-Max spent more than 80\% of their inference budget on non-essential steps (Redundancy Ratio $R=86.5\%$). Overall, these findings showed that redundancy was pervasive but varied substantially across LRMs.

\paragraph{Topological Pathologies: Diagnosing Distinct Modes of Redundancy.}
Our graph-topological metrics suggested that redundancy was not monolithic; instead, it appeared through distinct failure patterns across architectures. \textbf{(1) Cyclic Complexity}: DeepSeek-R1 and its distilled variant (70B) exhibited a non-linear style with frequent recurrence, showing high Average Degree ($\overline{D}\approx 1.75$ and $1.96$) and pronounced \textit{Logical Epicenters} (Max In-degree $\approx 6.8$ and $6.3$). This pattern indicated that computation was disproportionately spent around a small number of hub nodes where the model repeatedly branched or looped back. \textbf{(2) Semantic Verbosity}: In contrast, Qwen3-Max combined extreme uncertainty ($U=0.955$), a high isolated node ratio (10.5\%), and a high self-loop ratio (15.0\%) with the lowest presence of logical epicenters. This pattern pointed to self-clarification and verbosity as primary drivers of redundancy, manifesting as global looseness rather than the local congestion observed in DeepSeek. \textbf{(3) Local Over-Optimization}: Gemini-3-Pro maintained a largely linear structure ($\overline{D}\approx1.13$) but showed high backtracking ratios, suggesting a concise backbone with micro-inefficiencies rather than structural sprawl.

\paragraph{The Pareto Frontier and Inherited Distillation Bloat.}
Within this framework, \textbf{gpt-oss-120b defines the Pareto frontier}, achieving $>80\%$ accuracy while maintaining low redundancy ($R$). In contrast, distilled LRMs consistently exhibited $R$ values exceeding 69\%, with DeepSeek-R1 reaching 78.0\% (the second highest). This pattern suggests that distillation not only transfers knowledge but also inherits redundancy from the teacher model, potentially amplifying structural bloat when compressed into smaller architectures.

\begin{table*}[t]
\centering
\caption{Evaluation results (best/worst marked) for 21 LRMs. All metrics were averaged over the evaluated CoTs. Abbreviations: isolated node ratio ($Iso$), maximum in-degree ($ind_{\max}$), maximum out-degree ($out_{\max}$), self-loop ratio ($Self$), backward ratio ($Back$), and uncertainty ratio ($Unc$). Note: $\uparrow$ indicates higher is better and $\downarrow$ indicates lower is better. \textbf{Bold} marks the highest efficiency, and \underline{underline} marks the lowest efficiency.}

\scriptsize
\setlength{\tabcolsep}{2.22pt} 
\begin{tabular}{l|cc|cccccccc|ccc}
\toprule
\textbf{Model Name} & \multicolumn{2}{c|}{\textbf{Basic Stats}} & \multicolumn{8}{c|}{\textbf{Graph Topological Metrics}} & \multicolumn{3}{c}{\textbf{Core Efficiency}} \\
& $Acc$ $\uparrow$ & $Tokens$ $\downarrow$ & $|V|$ $\downarrow$ & $Iso$ $\downarrow$ & $|E|$ $\downarrow$ & $\overline D$ $\downarrow$ & $in_{max}$ $\downarrow$ & $out_{max}$ $\downarrow$ & $Self$ $\downarrow$ & $Back$ $\downarrow$ & $L_{eff}$ $\downarrow$ & $R$ $\downarrow$ & $Unc$ $\downarrow$ \\
\midrule
\rowcolor[HTML]{F2F2F2} \multicolumn{14}{l}{\textbf{Proprietary Models}} \\
Claude-Sonnet-4.5 & 0.775 & \textbf{450.5} & 21.8 & 0.012  & 24.0 & 1.138 & 2.71 & 2.59 & 0.087 & 0.082 & 11.9 & 0.368 & \textbf{0.031}\\
Doubao-Seed-1.6 & 0.855 & 1462.3 & 26.0 & \textbf{0.003} & 41.7 & 1.661 & 3.72 & 3.54 & 0.102 & 0.074 & 9.6 & 0.502 & 0.171 \\
Doubao-Seed-1.6-Flash-Thinking & 0.803 & 2756.0 & 35.8 & 0.004 & 59.9 & 1.761 & 4.41 & 4.21 & 0.123 & 0.072 & 9.8 & 0.549 & 0.241 \\
Gemini-2.5-Flash-Thinking & 0.797 & 5146.6 & 13.4 & 0.027 & 15.7 & 1.230 & 3.10 & 2.80 & \textbf{0.033} & 0.077 & 8.9 & 0.377 & 0.037 \\
Gemini-2.5-Pro & 0.789 & 3557.0 & 13.0 & 0.032 & 14.8 & 1.229 & 3.00 & 2.90 & 0.050 & 0.069 & 8.9 & 0.393 & 0.073 \\
Gemini-3-Pro & 0.703 & 4160.4 & \textbf{10.7} & 0.040 & \textbf{11.2} & \textbf{1.129} & 2.06 & 1.94 & 0.050 & \underline{0.153} & \textbf{7.0} & \textbf{0.333} & 0.049 \\
Qwen3-Max & 0.853 & 6954.8 & 181.2 & \underline{0.105} & 265.0 & 1.617 & \textbf{1.59} & \textbf{1.64} & \underline{0.150} & 0.123 & \underline{46.8} & \underline{0.865}  & \underline{0.955}\\

\midrule
\rowcolor[HTML]{F2F2F2} \multicolumn{14}{l}{\textbf{Open-Source Models}} \\
DeepSeek-R1 & 0.828 & 4313.4 & 149.8 & 0.030 & 233.0 & 1.749 & \underline{6.84} & \underline{6.70} & 0.112 & 0.070 & 12.4 & 0.780 & 0.636 \\
DeepSeek-V3.2 & 0.832 & 1794.3 & 45.9 & 0.068 & 63.9 & 1.484 & 3.10 & 2.95 & 0.098 & 0.075 & 10.3 & 0.580 & 0.382 \\
GLM-4.6 & 0.821 & 2673.6 & 61.8 & 0.048 & 101.7 & 1.668 & 5.37 & 5.00 & 0.112 & 0.059 & 8.8 & 0.667 & 0.646 \\
Kimi-K2-Thinking & 0.778 & 2474.4 & 42.1 & 0.031 & 69.1 & 1.784 & 5.74 & 5.40 & 0.149 & 0.060 & 8.6 & 0.687 & 0.561 \\
Qwen3-235B-A22B-Thinking-2507 & 0.868 & 5823.0 & 117.4 & 0.020 & 178.4 & 1.600 & 5.53 & 5.23 & 0.100 & 0.129 & 12.3 & 0.693 & 0.608 \\
Qwen3-30B-A3B-Thinking-2507 & \textbf{0.869} & 5077.3 & 93.1 & 0.019 & 145.7 & 1.656 & 5.67 & 5.34 & 0.114 & 0.132 & 12.0 & 0.688 & 0.684\\
gpt-oss-120b & 0.824 & 1080.1 & 21.5 & 0.033 & 28.2 & 1.331 & 2.65 & 2.51 & 0.088 & \textbf{0.044} & 9.1 & 0.336 & 0.164 \\
gpt-oss-20b & 0.758 & 2185.6 & 34.4 & 0.051 & 51.5 & 1.367 & 2.44 & 2.32 & 0.081 & 0.095 & 8.8 & 0.498 & 0.366 \\

\midrule
\rowcolor[HTML]{F2F2F2} \multicolumn{14}{l}{\textbf{Distilled Models}} \\
DeepSeek-R1-0528-Qwen3-8B & 0.808 & \underline{8816.8} & \underline{303.1} & 0.059 & \underline{449.1} & 1.677 & 6.21 & 6.00 & 0.139 & 0.118 & 12.3 & 0.742  & 0.820\\
DeepSeek-R1-Distill-Llama-70B & 0.796 & 3402.8 & 91.9 & 0.026 & 152.1 & \underline{1.964} & 6.31 & 6.31 & 0.147 & 0.082 & 10.1 & 0.725 & 0.694 \\
DeepSeek-R1-Distill-Llama-8B & 0.598 & 5090.6 & 159.2 & 0.025 & 225.7 & 1.583 & 4.90 & 4.81 & 0.117 & 0.108 & 12.2 & 0.753 & 0.626 \\
DeepSeek-R1-Distill-Qwen-14B & 0.717 & 3256.1 & 96.2 & 0.035 & 142.0 & 1.409 & 5.79 & 5.57 & 0.110 & 0.080 & 11.2 & 0.754 & 0.427 \\
DeepSeek-R1-Distill-Qwen-32B & 0.708 & 2141.0 & 63.5 & 0.038 & 91.9 & 1.275 & 4.34 & 4.20 & 0.081 & 0.059 & 11.1 & 0.699 & 0.282 \\
DeepSeek-R1-Distill-Qwen-7B & \underline{0.560} & 3596.5 & 108.7 & 0.038 & 153.2 & 1.442 & 4.77 & 4.53 & 0.118 & 0.083 & 12.9 & 0.710 & 0.512 \\
\bottomrule
\end{tabular}
\label{tab:benchmark_results}
\end{table*}

\subsection{Insights into CoT Reasoning}

\paragraph{Was Verbosity Used as a Compensation Strategy?} Figure~\ref{fig:token_distribution_comparison} illustrates that Flash and smaller-parameter models exhibit a distinct rightward shift and increased dispersion in token lengths. This pattern suggests a reliance on test-time scaling, wherein models generate additional tokens to compensate for limited per-step reasoning capabilities. Furthermore, while all models display a heavy right tail, Gemini-2.5-Flash-Thinking is notable for producing extreme outliers (exceeding 60{,}000 tokens), indicative of instability and ineffective halting mechanisms in edge cases.

\begin{wrapfigure}{r}{0.5\columnwidth}
  \centering
  \vspace{-0.5\baselineskip}
  \includegraphics[width=\linewidth, keepaspectratio]{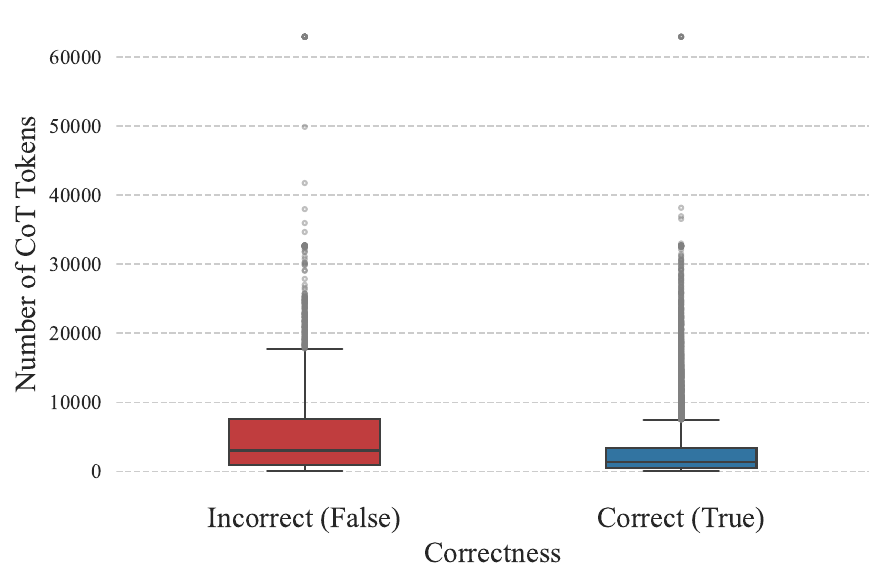}
  \caption{Distribution of CoT token counts for correct versus incorrect outcomes across evaluated models.}
    \vspace{-0.5\baselineskip} 
  \label{fig:tokens_boxplot_by_correctness}
\end{wrapfigure}

\paragraph{Did Failure Increase Verbosity?}
Figure~\ref{fig:tokens_boxplot_by_correctness} demonstrates a strong correlation between error rates and excessive generation. Incorrect responses exhibit higher median token counts and wider interquartile ranges, implying that models increase verbosity in an effort to recover from erroneous trajectories. Notably, the incorrect group features a dense cluster of outliers (10k--30k tokens), consistent with inefficient looping behavior. While extreme outliers ($>60$k) appear in both groups, indicating a systemic halting deficiency, this uncontrolled redundancy is significantly exacerbated during reasoning breakdowns.

\begin{figure*}[t]
  \includegraphics[width=0.96\linewidth]{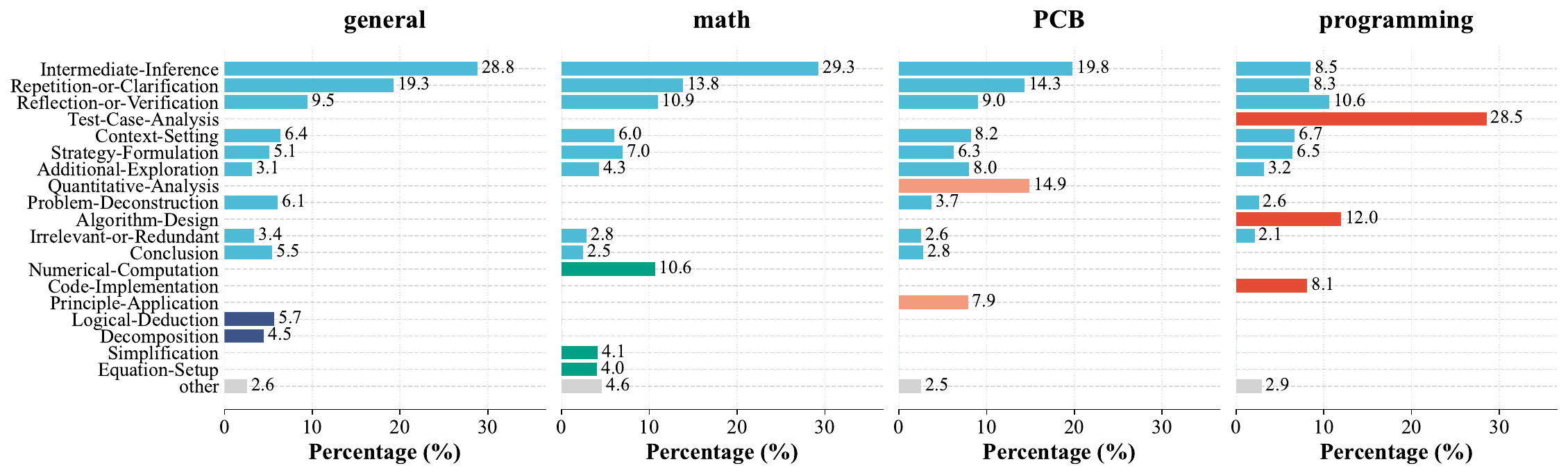}
  \caption{Functional role distribution of CoT steps across four domains (General Reasoning, Math, Programming, and PCB). Each chart shows the proportions of universal and domain-specific reasoning roles, highlighting shared structure and domain-adaptive patterns in LRMs.}
  \label{fig:category_distribution}
\end{figure*}
\begin{figure*}[t]
  \includegraphics[width=0.98\linewidth]{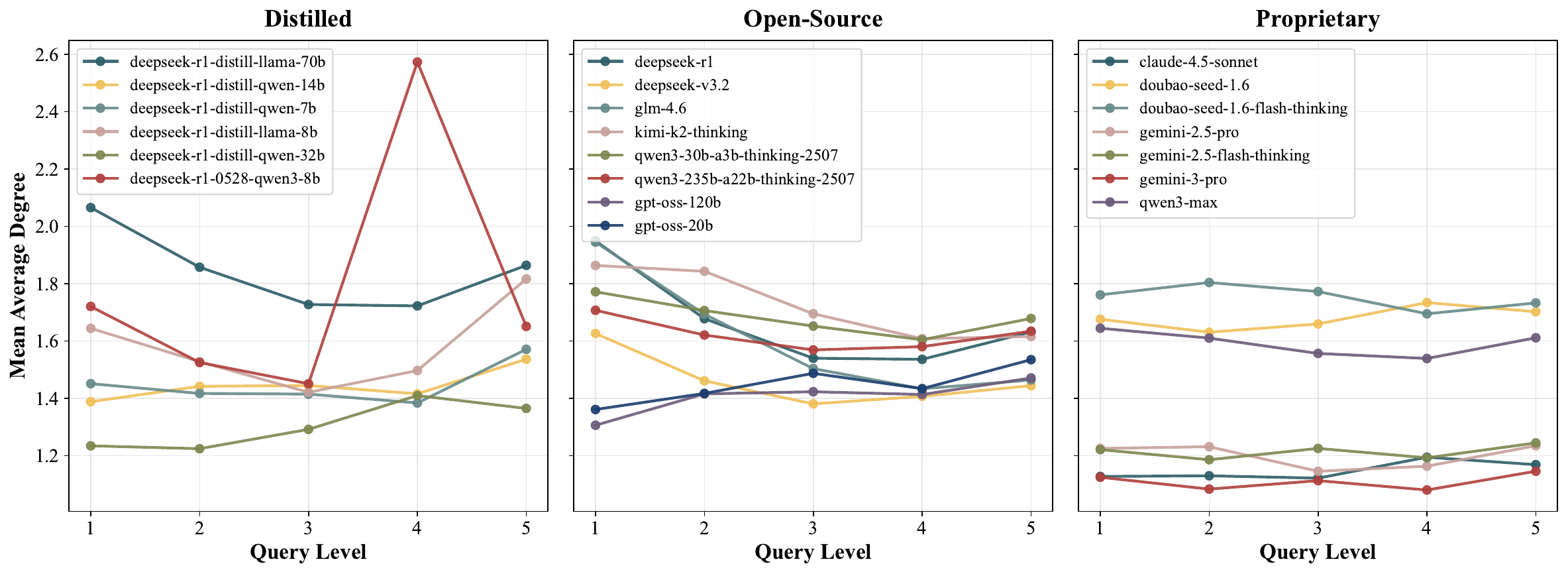}   
  \caption {Average degree ($\overline D$) of CoT graphs across query levels, grouped by model family (distilled, open-source, proprietary). }  \label{fig:avg_degree_vs_query_level_by_model_type}
\end{figure*}
\paragraph{Did Reasoning Patterns Generalize Across Domains?}
Figure~\ref{fig:category_distribution} suggested a dual structure in LRM reasoning: a shared logical backbone alongside domain-specific strategies. Across all domains, \textit{Intermediate-Inference} and \textit{Reflection-or-Verification} consistently formed a derivation--verification core. Notably, a stable 10--20\% share of \textit{Repetition-or-Clarification} persisted across domains, indicating that models frequently relied on redundancy to maintain context. Domain adaptations were also clear: Math emphasized formal computation, whereas PCB prioritized principle application. In \textbf{Programming}, reasoning shifted toward verification and was dominated by \textit{Test-Case-Analysis} (28.5\%), reflecting result-driven engineering logic rather than purely semantic generation.

\paragraph{Did Complexity Scale with Difficulty?}
Figure~\ref{fig:avg_degree_vs_query_level_by_model_type} reveals divergent topological behaviors across model families. Proprietary models, such as Gemini-3-Pro and Claude-Sonnet-4.5, maintained robust stability ($\overline{D}\approx 1.1\text{--}1.2$), characterized by largely linear reasoning independent of difficulty. In contrast, open-source models exhibited elevated baselines ($1.4\text{--}1.8$) and a distinct U-shaped trajectory: over-reasoning on simple queries, achieving peak efficiency (minimal $\overline{D}$) when difficulty matched model capability, and regressing into redundant backtracking when overwhelmed. Notably, DeepSeek-R1-0528-qwen-8B exhibited a sharp spike to $\overline{D}\approx 2.6$ at Level~4, signaling a topological collapse driven by excessive looping under high cognitive load.

\paragraph{Where Did Redundancy Accumulate?} Figure~\ref{fig:redundant_position_kde} reveals a consistent temporal pattern across models: redundancy is initially low, settles into a prolonged mid-stage plateau, and rises sharply towards the end. This trajectory suggests that mid-stage redundancy plays a functional role in maintaining context and stabilizing reasoning, rendering it difficult to excise without degrading performance—a significant challenge for CoT compression. Additionally, several models exhibit a distinct pre-answer peak (0.8--0.95), indicative of intensive self-checking and summarization for confidence calibration. Overall, redundancy appears not as sporadic noise, but as a structural mechanism essential for stability and assurance.

\begin{figure}[t]
  \centering
  \includegraphics[width=0.75\columnwidth]{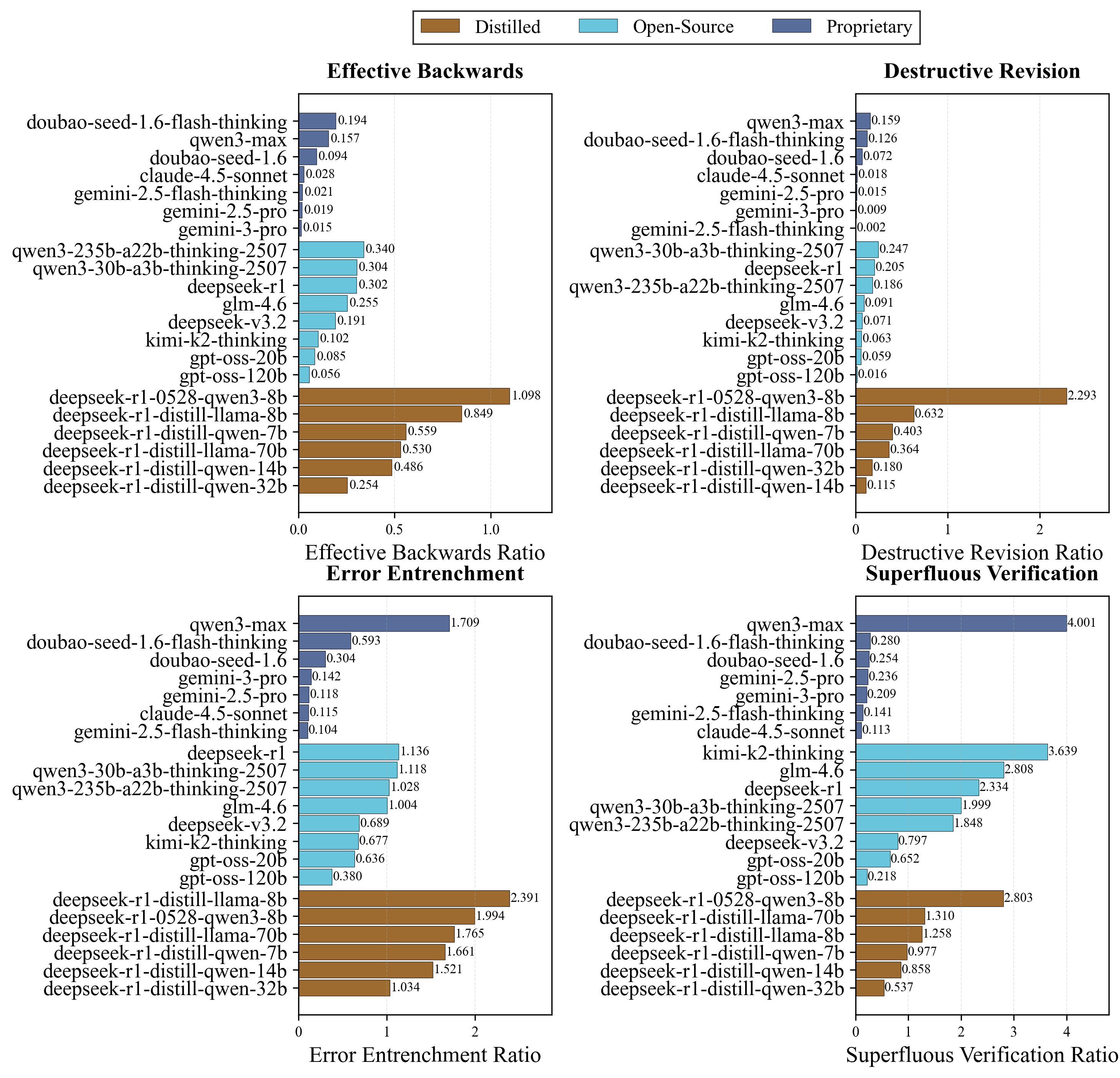}
  \caption{Distribution of external redundancy patterns. The bar charts show the fraction of reasoning steps involved in four types of answer transitions, highlighting varying degrees of post-answer redundancy.}
  \label{fig:pattern_analysis_4_subplots}
\end{figure}

\paragraph{Was Post-Answer Reasoning Beneficial?}Motivated by the prevalently high uncertainty ratios ($>40\%$) observed in Section~\ref{subsec:overall_performance}, we examined \textit{external redundancy}, defined as reasoning generated subsequent to the initial answer \citep{hong2025reconsideringoverthinkingpenalizinginternal}. We classified backward reasoning steps into four transition modes ($Answer_{n}\to Answer_{n+1}$): \textbf{Destructive Revision (DR, Correct $\to$ Incorrect)}, where instability causes a correct solution to be discarded; \textbf{Superfluous Verification (SV, Correct $\to$ Correct)}, representing redundant checking after achieving accuracy; \textbf{Error Entrenchment (EE, Incorrect $\to$ Incorrect)}, characterizing unproductive looping among errors; and \textbf{Effective Backwards (EB, Incorrect $\to$ Correct)}, reflecting successful, albeit inefficient, self-correction.

Figure~\ref{fig:pattern_analysis_4_subplots} reveals significant stability disparities. Distilled models exhibited disproportionately high \textbf{DR}, with DeepSeek-R1-0528-Qwen3-8B exceeding a 2.0 ratio. This suggests that such models mimic the \emph{form} of reflection without the requisite stability, often degrading correct initial derivations. They also displayed high \textbf{EB}, indicating a reliance on stochastic trial-and-error. In contrast, proprietary models (e.g., Gemini-3-Pro) demonstrated robust convergence, maintaining low \textbf{DR} and \textbf{EB} by locking onto correct paths early. Except for the Gemini series, Claude-Sonnet-4.5, Doubao-Seed-1.6, and gpt-oss-120b, \textbf{EE} values remained generally high ($>1.0$), highlighting a systematic struggle to recover from logical failures. Finally, Qwen3-Max and Kimi-K2-Thinking displayed extreme \textbf{SV} (3--4 orders of magnitude higher), indicative of post-solution looping that degrades latency without accuracy gains. Detailed case studies are provided in Appendix~\ref{sec:case_studies}. 

These findings imply that current distillation recipes often transfer the \textit{surface-level verbosity} of reasoning without the corresponding \textit{verification capabilities}, resulting in a ``reasoning illusion'' where longer generations degrade rather than improve reliability.

\section{Conclusion}
We introduced \textbf{CoTJudger}, an automated and general framework that transforms free-form CoT into directed dependency graphs to assess reasoning efficiency via validity-aware path analysis. By disentangling the Shortest Effective Path from structurally redundant segments, CoTJudger transcends token-length proxies, enabling an interpretable, topology-based diagnosis of over-reasoning. Across large-scale evaluations of 21 LRMs spanning proprietary, open-source, and distilled paradigms, we found that redundancy is pervasive yet highly model-dependent: while some models expand computation far beyond necessity, others maintain near-linear reasoning but still exhibit localized backtracking and post-answer looping. Our analysis further demonstrates that inefficiency is not a monolithic phenomenon but emerges through distinct patterns, such as cyclic recurrence, semantic verbosity, and unstable revisions, that can be localized across domains and difficulty levels. Collectively, these results redefine reasoning quality to encompass not just correctness but the structural necessity of the reasoning trajectory, providing actionable signals for developing LRMs that are both accurate and efficient.

\clearpage

\bibliographystyle{plainnat}
\bibliography{main}

\clearpage

\beginappendix

\section{Dataset}
\label{sec:query_details}
\noindent \textbf{Data Source.} Query dataset is sourced from publicly available benchmarks: OmniMath \citep{gao2024omnimathuniversalolympiadlevel} (Math), HumanEval \citep{chen2021codex} (Programming), GPQA \citep{rein2023gpqagraduatelevelgoogleproofqa} (PCB), and Big-Bench Hard \citep{suzgun2022challengingbigbenchtaskschainofthought} (General Reasoning). 

\noindent \textbf{Data Cleaning and Adaptation.} To adapt to the subsequent fully automated CoT analysis and correctness verification pipeline, we conducted critical cleaning on the initial samples:

\begin{itemize}

\item Removal of queries with excessively long answers to avoid introducing noise due to differences in text generation during evaluation.

\item Manual exclusion of queries with open-ended answers or requiring free-form responses to ensure that all problems have clear, automatically verifiable ground-truth answers.

\item Special handling for programming queries: Since their correctness can be objectively judged through code execution, the above two cleaning steps are not applicable. We retain the original queries and strictly follow the execution environment of the source dataset for verification.
\end{itemize}

\noindent \textbf{Subdomain Distribution.} The subdomain distribution of query dataset is shown in Table~\ref{tab:domain_distribution}.

\section{Framework Development and Validation on Development Set}

\label{sec:development}

We utilized this development set (2,688 CoTs from Gemini-3-Pro, Qwen3-30B-A3B-Thinking-2507 and DeepSeek-R1-Llama-70B) for multiple rounds of modular design and validation for the framework's iteration.

\noindent \textbf{Pipeline Design and Tuning.} Each step of the entire evaluation pipeline, such as heuristic rules for step segmentation, boundary definitions for classification, and reconnection logic for graph construction, was tested and optimized through case analysis on the development set. We continuously adjusted prompts, code designs, and algorithm parameters by observing failure cases to ensure the pipeline’s robustness to complex and diverse CoTs.

\noindent \textbf{Taxonomy Construction.} Our CoT node classification system was not pre-defined but systematically constructed through multiple rounds of sampling, open-ended induction, cross-domain integration, and manual refinement of a large number of CoT steps in the development set. This ensures its empirical foundation and completeness.

\noindent \textbf{Framework Validation Experiments.} We ran the complete evaluation pipeline on this development set, conducting case analysis and validation experiments to verify the rationality of the pipeline.

To validate the stability and effectiveness of the pipeline, we executed the full pipeline three times on the core development set. First, we recorded the output variations of the Step Segmentation, Node Classification, and Answer Verification modules to calculate their consistency. Next, to assess the validity of the methodology proposed by CoTJudger, we designed experiments to examine whether the Shortest Effective Paths (SEPs) derived from Path Extraction truly retain the core logic. We employed GPT-5 to rewrite the SEPs extracted from the CoTs generated across the three runs. This rewriting process preserved the original path logic without adding new content, focusing solely on logical organization and expression smoothing, while masking the final answer. The rewritten content was then fed back into the original subject LLM for re-inferring to evaluate whether the answer accuracy was maintained.

\begin{table*}[t]
\centering
\small 
\caption{Detailed statistics of the dataset distribution across four main domains and their subdomains. The percentages in parentheses indicate the relative weight of each subdomain within its parent domain.}
\label{tab:domain_distribution}
\begin{tabularx}{\textwidth}{l l X}
\toprule
\textbf{Domain} & \textbf{Count} & \textbf{Subdomains Distribution} \\
\midrule
\textbf{Math} & 364 & 
Logic (13.8\%), Algebra (13.1\%), Calculus (12.0\%), Linear Algebra (11.9\%), Probability (11.8\%), Geometry (12.1\%), Combinatorics (11.3\%), Number Theory (7.0\%), Math Puzzle (7.0\%) \\
\addlinespace 

\textbf{General Reasoning} & 270 & 
\textbf{11.1\% each:} Logical Deduction, Tracking Shuffled Objects; \newline
\textbf{3.7\% each:} Boolean Expressions, Causal Judgement, Date Understanding, Disambiguation QA, Dyck Languages, Formal Fallacies, Geometric Shapes, Hyperbaton, Movie Recommendation, Multistep Arithmetic Two, Navigate, Object Counting, Penguins in a Table, Reasoning about Colored Objects, Ruin Names, Salient Translation Error Detection, Snarks, Sports Understanding, Temporal Sequences, Web of Lies, Word Sorting \\
\addlinespace

\textbf{PCB} & 98 & 
Biology (37.8\%), Chemistry (31.6\%), Physics (30.6\%) \\
\addlinespace

\textbf{Programming} & 164 & 
String (22.6\%), List/Array (18.9\%), Math (13.4\%), Hash Map (9.7\%), DP (6.9\%), Simulation (6.1\%), Graph (5.4\%), Sort/Search (4.9\%), Stack/Queue (3.7\%), Linked List (3.0\%), Tree Binary (3.0\%), Bit Manipulation (2.4\%) \\
\bottomrule
\end{tabularx}
\end{table*}

The overall experimental results are presented in Table~\ref{tab:valid_exp}. The output consistency for each module exceeded 92\%, and the overall consistency across the three pipeline executions reached 94.2\%, demonstrating robust pipeline stability. Furthermore, the average answer retention rate of 95.1\% confirms that CoTJudger's path extraction strategy effectively preserves the core logic of the CoT.

\begin{table}[t]
    \centering
    \small
    \caption{Validation results for pipeline module stability and path effectiveness.}
    \label{tab:valid_exp}
    \begin{tabularx}{\linewidth}{l c X}
        \toprule
        \textbf{Module} & \textbf{Metric} & \textbf{Metric Description} \\
        \midrule
        Step Atomization & 94.1\% &
        Stability of step counts across runs, measured as $1 - CV$ where $CV=\frac{\sigma}{\mu}$. \\
        
        Classification & 92.1\% &
        Micro-F1 score of predicted labels,
        $F1_{\text{micro}}=\frac{2\sum TP}{2\sum TP+\sum FP+\sum FN}$. \\
        
        Answer Detection & 96.3\% &
        Stability of detected answer counts across runs, measured as $1 - CV$. \\
        
        Path Extraction \\
        and Validation & 95.1\% &
        Accuracy retention rate after rewriting,
        computed as $\frac{1}{N}\sum \mathbf{1}[y_{\text{new}} = y_{\text{original}}]$. \\
        \bottomrule
    \end{tabularx}
\end{table}


        
    

\section{Classification System}
\label{sec:appendix_taxonomy}

To ensure clarity and domain adaptability, the classification system was constructed through a four-stage iterative process based on the core development set:

\begin{enumerate} \item \textbf{Sampling and Annotation:} We conducted stratified random sampling from 2,688 CoTs generated by a diverse set of models (Gemini-3-Pro, Qwen3-30B-A3B-Thinking-2507, and DeepSeek-R1-Llama-70B) to ensure coverage of varied reasoning styles. \item \textbf{Domain-Specific Induction:} For each domain, LLMs were employed to perform initial clustering and open-coding of node functions, inducing a preliminary category list specific to that field. \item \textbf{Cross-Domain Aggregation:} We merged the domain-specific lists to identify overlapping functions. Universal core categories (e.g., \textit{Problem-Deconstruction}, \textit{Context-Setting}) were extracted to form the top tier, while domain-unique actions were retained in the second tier. \item \textbf{System Refinement:} The aggregated system underwent multiple rounds of manual review to resolve ambiguities. Special attention was paid to boundary cases, such as establishing strict criteria to distinguish between \textit{Reflection-or-Verification} (evaluative) and \textit{Correction-or-Refinement} (executive). \end{enumerate}

The final complete taxonomy is provided in Table~\ref{tab:full_taxonomy}.

\begin{table*}[t]
\centering
\small
\caption{The proposed two-tier classification taxonomy.}
\label{tab:full_taxonomy}
\begin{tabularx}{\textwidth}{l l >{\hsize=0.7\hsize}X >{\hsize=1.3\hsize}X}
\toprule
\textbf{Domain} & \textbf{Category} & \textbf{Sub-categories} & \textbf{Definition} \\
\midrule
\multicolumn{4}{c}{\textit{\textbf{Universal Classification System}}} \\
\midrule
\multirow{10}{*}{Universal} 
& Problem-Deconstruction & Problem-Understanding, Info-Extraction, Info-Organization & Analyzes prompt to extract, parse, or organize key info, constraints, and core questions. \\
& Context-Setting & Definition, Assumption, Concept-Explanation & Defines terms, states assumptions, or explains core concepts required for reasoning. \\
& Strategy-Formulation & – & Outlines a high-level plan, sequence of steps, or specific method to solve the problem. \\
& Intermediate-Inference & – & Makes a deductive step or logical implication based on current info to advance the solution state. \\
& Reflection-or-Verification & Verification, Reflection, Premise-Reassessment, Hypothesis-Rejection & Assesses previous steps for correctness or consistency without taking new action. \\
& Correction-or-Refinement & Correction, Refinement, Simplification, Adjustment & \textbf{Actively} implements a fix or adjustment based on new insights or errors in a previous step. \\
& Repetition-or-Clarification & – & Reiterates previous content or adds inconsequential supplements without changing core meaning. \\
& Additional-Exploration & – & Investigates alternative paths or assumptions when the current one is viable but not definitive. \\
& Irrelevant-or-Redundant & – & Steps that are verbose, off-topic, or do not contribute to the solution trajectory. \\
& Conclusion & – & Summarizes findings or provides the final answer. \\

\midrule
\multicolumn{4}{c}{\textit{\textbf{Domain-Specific Extensions}}} \\
\midrule
\multirow{5}{*}{Math} 
& Equation-Setup & – & Translates constraints into mathematical equations or expressions. \\
& Formula-Application & – & Selects and applies specific formulas, theorems, or rules. \\
& Simplification & – & Performs algebraic manipulation to simplify expressions. \\
& Numerical-Computation & – & Performs arithmetic calculations to arrive at numerical values. \\
& Proof-Step & – & Formal steps in a logical proof (lemma, inference rule, case establishment). \\
\addlinespace

\multirow{4}{*}{\shortstack[l]{General\\Reasoning}} 
& Decomposition & – & Breaks complex tasks into manageable sub-goals. \\
& Logical-Deduction & – & Applies logic rules (e.g., if A then B) to derive conclusions. \\
& Strategic-Assessment & – & Evaluates multiple potential paths to determine the best one. \\
& Pattern-Recognition & – & Identifies recurring trends or sequences to guide the solution. \\
\addlinespace

\multirow{3}{*}{PCB} 
& Principle-Application & – & Applies scientific principles (e.g., F=ma, Ohm's Law). \\
& Comparative-Analysis & – & Compares scenarios or models to draw insights. \\
& Quantitative-Analysis & – & Performs domain-specific calculations or unit conversions. \\
\addlinespace

\multirow{5}{*}{Programming} 
& Algorithm-Design & – & Describes logic/plans using pseudocode or data structures. \\
& Code-Implementation & – & Translates logic into concrete programming language syntax. \\
& Test-Case-Analysis & – & Verifies logic using specific inputs/outputs or traces. \\
& Optimization & – & Proposes changes to improve efficiency (time/space complexity). \\
& Debugging & – & Identifies errors in existing code and proposes fixes. \\
\bottomrule
\end{tabularx}
\end{table*}

\section{Prompts}
\label{sec:prompt}
To ensure the reliability of most automated modules that call LLM (GPT-5) , we employ Pydantic-based structured outputs to enforce predefined schema-conformance and stable data parsing.
\noindent \textbf{Step Atomization Prompt.}
After performing initial heuristic segmentation, we convert the original CoT into a list of splits. Subsequently, we further utilize LLM to perform merging and subsequent splitting operations on the elements in this list. The prompt is shown in Figure~\ref{fig:step_prompt}. To maintain the integrity of the original model output, the LLM is instructed to output the mapping of step IDs (e.g., \textit{``Merge steps i and j'', ``Split step i into j and j''}) instead of rewriting the entire CoT. This prevents the model introducing errors or biases during paraphrasing while leveraging its strong semantic understanding capability to ensure that each final node represents an atomic reasoning action. 

\begin{figure*}[htbp] \centering 
\begin{tcolorbox}[colback=gray!5, colframe=gray!50, title=\textbf{Input Prompt}] 
\small 
\textbf{[Instruction]} \ The following text is a CoT (Chain of Thought), which has been automatically split into basic splits. Most splits are already atomic reasoning steps, but some may need to be merged or split to create proper atomic steps.

\textbf{Splits of the CoT are as follows}: \ \textit{{Input sequence of splits, formatted as ``split ID: content''}}

\textbf{Task}: Identify splits that need special processing (merging or splitting). For splits that don't need changes, leave them as-is.

\textbf{Guidelines}: \begin{enumerate} \item \textbf{MERGING}: Combine consecutive splits that belong to the same logically independent step.  \item \textbf{SPLITTING}: If a single split contains multiple distinct atomic reasoning steps, split it into separate steps. Eg: \begin{itemize} \item a split has both reasoning and a separate answer/conclusion; \item a split involves both reasoning and subsequent twists or additional thinking..., etc. \end{itemize} \item \textbf{ATOMIC STEPS}: Each final step should represent one atomic reasoning operation or conclusion. \end{enumerate}

\textbf{[Domain Specific Instructions]} \ \textit{(If Domain == Programming)}: Code blocks/splits within ``` ``` are \textbf{already} atomic steps, \textbf{DON'T} merge or split it.

\hrulefill

\textbf{[Output Schema Constraint]} \ The model is constrained to output a JSON object adhering to the following structure:

\begin{verbatim} 
    { 
    "merge_operations": [ { 
        "split_ids": [integer, ...], // List of split IDs to merge 
        "summary": string // Summary of the merged step 
        } ], 
    "split_operations": [ { 
        "split_id": integer, // The split ID to split 
        "steps": [ { 
            "content": string, // The content for this sub-step 
            "summary": string // Summary of this sub-step 
            } ] 
        } ] 
    } 
\end{verbatim} 
\end{tcolorbox} 
\caption{The prompt and structured output schema used for refining step atomization.} 
\label{fig:step_prompt} 
\end{figure*}

\noindent \textbf{Answer Verification Prompt.}
To ensure rigorous evaluation across diverse domains, we adopt two distinct prompting strategies. For tasks of Math, General Reasoning and PCB, GPT5 is instructed to simultaneously locate the answer and verify its correctness by cross-referencing the logical consistency and the provided query and ground truth (Figure~\ref{fig:answer_verification_prompt}). Conversely, for the Programming domain, the prompt is restricted to answer location and extraction only (Figure~\ref{fig:code_location_prompt}). In this case, the model identifies the code solution, while the actual correctness verification is delegated to a deterministic code execution sandbox.
\begin{figure*}[htbp]
\centering
\begin{tcolorbox}[colback=gray!5, colframe=gray!50, title=\textbf{Prompt for Answer Verification (Math, PCB, Reasoning)}]
\small
\textbf{[Instruction]} \\
Please analyze the following CoT reasoning steps (domain: \textit{\{domain\}}) and identify steps containing final answers or conclusions to the problem:

\textbf{Original Question}: \textit{\{question\_text\}} \\
\textbf{Ground Truth Answer}: \textit{\{ground\_truth\_answer\}} \\
\textbf{CoT Reasoning Steps}: \textit{\{step\_id: content, ...\}}

\textbf{Task}:
\begin{enumerate}
    \item Identify steps providing a complete answer or conclusion to the problem (even if derived in advance).
    \item Extract the specific answer text content.
    \item Determine the correctness of the extracted answers by analyzing the requirements of the original question, ground truth, and the reasoning content.
    \item Provide clear reasoning explanations.
\end{enumerate}

\textbf{Note}:
\begin{itemize}
    \item \textbf{Intermediate/partial} results are \textbf{NOT} considered "answers".
    \item Extracted answers must exactly match the target required by the original question (e.g., if asking for "2x", "x" is invalid).
    \item Correctness is determined by cross-referencing the Question, Ground Truth, and Step Content.
\end{itemize}

\hrulefill

\textbf{[Output Schema Constraint]} \\
The model output must adhere to the following JSON structure:
\begin{verbatim}
{
  "step_answers": [
    {
      "step_id": integer,
      "contains_answer": boolean,   // Does this step contain a final answer?
      "answer_text": string,        // Extracted answer content
      "is_correct": boolean,        // Is the extracted answer correct?
      "reasoning": string           // Reasoning for detection and verification
    }
  ],
  "final_answer": string,           // The final answer text of the CoT
  "is_final_answer_correct": boolean // Overall correctness of the final answer
}
\end{verbatim}
\end{tcolorbox}
\caption{The prompt used for answer detection and logical verification in non-coding domains.}
\label{fig:answer_verification_prompt}
\end{figure*}

\begin{figure*}[htbp]
\centering
\begin{tcolorbox}[colback=gray!5, colframe=gray!50, title=\textbf{Prompt for Code Answer Location (Programming)}]
\small
\textbf{[Instruction]} \\
Please analyze the following CoT reasoning steps (domain: programming) and identify steps containing code answers that intend to solve the problem completely:

\textbf{Original Question Prompt}: \textit{\{question\_text\}} \\
\textbf{CoT Reasoning Steps}: \textit{\{step\_id: content, ...\}}

\textbf{Task}:
For each step containing a complete code answer to the problem:
\begin{enumerate}
    \item Provide simple reasoning for your detection.
    \item Extract the specific answer text (code implementation).
\end{enumerate}

\textbf{Note}: Focus on identifying and extracting code implementations that solve the problem completely. \textbf{Do NOT verify correctness yet.}

\hrulefill

\textbf{[Output Schema Constraint]} \\
The model output must adhere to the following JSON structure:
\begin{verbatim}
{
  "step_answers": [
    {
      "step_id": integer,
      "contains_answer": boolean, // Does this step contain a code solution?
      "answer_text": string,      // Extracted code implementation
      "reasoning": string         // Reasoning for detection
    }
  ],
  "final_answer": string          // The final extracted code solution
}
\end{verbatim}
\end{tcolorbox}
\caption{The prompt used for locating code solutions in Programming tasks. Note that correctness fields are omitted as verification is performed via code execution.}
\label{fig:code_location_prompt}
\end{figure*}

\noindent \textbf{Classification Prompt.}
To ensuring precise functional attribution, the LLM is provided with the full definition of our two-tier taxonomy (customized for the specific domain), the original problem context, and the segmented CoT steps. The prompt (Figure~\ref{fig:classification_prompt}) emphasizes the importance of analyzing the functional role of each step within the overall reasoning structure—rather than relying solely on surface content.

\begin{figure*}[htbp]
\centering
\begin{tcolorbox}[colback=gray!5, colframe=gray!50, title=\textbf{Prompt for Automated Node Classification}]
\small
\textbf{[Instruction]} \\
You are an expert in analyzing chain-of-thought reasoning patterns in \textit{\{domain\}}.

Please classify each reasoning step according to its function in the CoT structure, referring to the categories in the classification list and the original question.

\textbf{Original Question}: \textit{\{question\_text\}}

\textbf{Classification System}: \\
\textit{\{List of categories and definitions derived from the Two-Tier Taxonomy\}}

\textbf{CoT Reasoning Steps to Classify}: \\
\textit{\{step\_id: content, ...\}}

\textbf{Requirements}:
\begin{enumerate}
    \item Provide clear reasoning for each classification decision, after analyzing the step's content or function, categories' meaning in the list and the original question.
    \item Assign each step to exactly one category from the available options above that best describes its reasoning function.
    \item Focus on the step's reasoning function in the CoT structure rather than just its content.
    \item \textbf{Every step} (\textit{\{step\_ids\_range\}}) should be assigned a category; do not make omissions or misplacements.
\end{enumerate}

\hrulefill

\textbf{[Output Schema Constraint]} \\
The model output must adhere to the following JSON structure:
\begin{verbatim}
{
  "reasoning": string, // Global reasoning or analysis for the batch
  "classifications": [
    {
      "step_id": integer,
      "category": string,  // The assigned category name
      "reasoning": string  // Specific analysis of why this step fits the category
    }
  ]
}
\end{verbatim}
\end{tcolorbox}
\caption{The prompt used to classify CoT nodes into the standardized taxonomy. The \textit{Classification System} section is dynamically populated with the relevant domain-specific definitions.}
\label{fig:classification_prompt}
\end{figure*}

\noindent \textbf{Graph Building Prompt.}
We employ two distinct LLM-based strategies to establish edge connections: (1) Highly semantic Overlap Detection (Figure~\ref{fig:repetition_prompt}), which identifies nodes that repeat or reclarify previous content to assign same node id and construct self-loop edge, and (2) Target Dependency Identification (Figure~\ref{fig:target_identification_prompt}), which determines the specific predecessor node targeted by relational operations (e.g., \textit{Reflection-or-Verification}, \textit{Correction-or-Refinement}) to form directed functional edges. For \textit{Reflection-or-Verification}, GPT-5 is also instructed to analyze whether the node affirms or negates its target node.

\begin{figure*}[htbp]
\centering
\begin{tcolorbox}[colback=gray!5, colframe=gray!50, title=\textbf{Prompt for Repetition/Redundancy Detection}]
\small
\textbf{[Instruction]} \\
You are an expert in analyzing chain-of-thought reasoning patterns. Given an original question and its reasoning steps, identify redundant relationships between steps.

\textbf{Original Question}: \textit{\{query\_text\}}

\textbf{Task}: Analyze the following chain-of-thought reasoning steps to identify pairs where a step is a redundant addition of a certain earlier step. A step can be marked as a redundant addition if it satisfies:
\begin{itemize}
    \item Essentially \textbf{semantically reiterating or expanding content} from previous steps without adding key meaning.
    \item Minor modifications or inconsequential supplementations of \textbf{one} previous step without changing its core logic.
\end{itemize}

\textbf{CoT Steps}: \textit{\{step\_id: content (Category), ...\}}

\textbf{Specific Requirements}:
\begin{enumerate}
    \item \textbf{Look for Semantic Similarity}: Don't just look for exact wording. If a step merely supplements a previous step without advancing core logic, it counts.
    \item \textbf{Check Logic Core}: If the step produces a new intermediate result, it is likely NOT redundant.
    \item \textbf{Category Hint}: Pay special attention to steps marked "Repetition-or-Clarification".
    \item \textbf{One-to-One Mapping}: A step repeats or clarifies at most one specific previous step.
\end{enumerate}

\hrulefill

\textbf{[Output Schema Constraint]} \\
The model output must adhere to the following JSON structure:
\begin{verbatim}
{
  "contains_repetition": boolean,
  "pairs": [
    {
      "step_id_1": integer, // The earlier step being repeated
      "step_id_2": integer, // The current step acting as repetition
      "reasoning": string   // Explanation of redundancy
    }
  ]
}
\end{verbatim}
\end{tcolorbox}
\caption{The prompt used to detect semantic overlaps.}
\label{fig:repetition_prompt}
\end{figure*}

\begin{figure*}[htbp]
\centering
\begin{tcolorbox}[colback=gray!5, colframe=gray!50, title=\textbf{Prompt for Relational Target Identification}]
\small
\textbf{[Instruction]} \\
You are an expert in analyzing reasoning logic within LLM Chain-of-Thought (CoT). Given an original question and its reasoning steps, analyze the logical dependencies and identify the target step for the current step.

\textbf{Original Question}: \textit{\{query\_text\}}

\textbf{Task Requirement}: \textit{\{Dynamic Task Prompt based on Category\}}
\begin{itemize}
    \item \textit{For Reflection}: Identify which \textbf{previous step} this current step is evaluating, checking, or verifying...
    \item \textit{For Correction}: Identify which \textbf{previous step} contains the specific content or error being corrected...
    \item \textit{For Exploration}: Identify which \textbf{previous step} established the method/assumption being shifted away from...
\end{itemize}

\textbf{All Steps}: \textit{\{step\_id: content (Category), ...\}} \\
\textbf{Current Step}: \textit{\{step\_id: content (Category)\}}

\textbf{Requirements}:
\begin{enumerate}
    \item \textbf{Accuracy}: Identify ONE predecessor step that best matches the task requirement.
    \item \textbf{Temporal Constraint}: The target step must precede the current step.
    \item \textbf{Logical Relevance}: The target must have logical relevance to the current step's function.
\end{enumerate}

\textit{[Conditional Instruction for Reflection Steps]} \\
Determine whether the current step affirms or negates the target step.

\hrulefill

\textbf{[Output Schema Constraint]} \\
The model output must adhere to the following JSON structure:
\begin{verbatim}
{
  "target_step": integer, // The ID of the identified predecessor
  "is_affirm": boolean,   // (Only for Reflection) True if affirming, False if negating
  "reasoning": string     // Logic for identification
}
\end{verbatim}
\end{tcolorbox}
\caption{The prompt used to identify the target node for relational categories (\textit{Reflection}, \textit{Correction}, \textit{Additional-Exploration}). The "Task Requirement" field is dynamically injected based on the current node's category rules.}
\label{fig:target_identification_prompt}
\end{figure*}

\noindent \textbf{Path Validation Prompt.}
Since Shortest Effective Paths (SEPs) from Path Extraction Module are formed by algorithmically concatenating non-adjacent nodes, they often lack grammatical fluency. Consequently, the prompt (Figure~\ref{fig:path_validation_prompt}) instructs the LLM to disregard surface-level disfluency and choppy transitions, focusing on whether the core logical flow remains intact and sufficient to derive the final answer.

\begin{figure*}[htbp]
\centering
\begin{tcolorbox}[colback=gray!5, colframe=gray!50, title=\textbf{Prompt for Path Logic Validation \& Extraction}]
\small
\textbf{[Role \& Context]} \\
You are a \textbf{Reasoning Content Analyst}. Your sole function is to evaluate the validation of a reasoning path extracted algorithmically from a longer Chain of Thought.

\textbf{CRITICAL INSTRUCTION}: Because this path is an algorithmic concatenation, it will likely be \textbf{disfluent, choppy, and grammatically awkward}. You must ignore these surface issues.

\textbf{Original Problem}: \textit{\{query\_text\}} \\
\textbf{Extracted Reasoning Path}: \textit{\{Concatenated steps of the extracted path\}}

\textbf{[Task]}
\begin{enumerate}
    \item \textbf{Determine Validity}: A path is "Valid" if the core content of the entire path essentially leads to its final answer.
    \item \textbf{Extract Core Logic}: Summarize the essential logical flow, focusing on the content pushing the deduction forward, while smoothing over the lack of transitions.
    \item \textbf{List Issues}: List the issues if the path is invalid.
\end{enumerate}

\hrulefill

\textbf{[Output Schema Constraint]} \\
The model output must adhere to the following JSON structure:
\begin{verbatim}
{
  "is_valid": boolean,        // Is the path logically sufficient to reach the answer?
  "extracted_logic": string,  // The smoothed, core logical flow of the path
  "issues": [string],         // List of specific issues (if any)
  "reasoning": string         // Explanation of the validation decision
}
\end{verbatim}
\end{tcolorbox}
\caption{The prompt used to validate extracted reasoning paths. It instructs the model to overlook algorithmic artifacts (disfluency) and verify if the semantic logic remains sufficient to solve the problem.}
\label{fig:path_validation_prompt}
\end{figure*}

\section{Experimental Setup}
\label{sec:setup}
\textbf{Inference Configuration.} For models accessed via API services, we used their default parameter configurations and enabled the official thinking mode, extracting CoTs from the `reasoning\_content` field of returned results. 

For models equipped with built-in thinking mechanisms (such as Qwen3-30B-A3B-Thinking-2507), we directly utilize their default generated content. For models that require specific formats to trigger thinking (such as DeepSeek-R1 and its distilled variants), we add the instruction "Please include your thinking process within <think> and </think>" to the prompt. Other models use only the query itself as the prompt.

\section{Cases of External Redundancy}
\label{sec:case_studies}
\noindent \textbf{A case of Deconstructive Revision.}
Here is a case of the Deconstructive Revision phenomenon of DeepSeek-R1-0528-Qwen3-8B (Figure~\ref{fig:DR}).

\begin{figure*}[htbp] \centering 
\begin{tcolorbox}[colback=gray!5, colframe=gray!50, title=\textbf{Deconstructive Revision}] 
\small 
\textbf{[Query]} \ Which of the following is a humorous edit of this artist or movie name: 'rain man'? Options: (A) ruin man (B) rains man (C) rain men (D) rainmman

\textbf{Query Domain}: Subset "ruin names" of General Reasoning 

\textbf{Query Level}: 2

\textbf{Ground Truth}: A

\hrulefill

\textbf{CoT} ...I recall that in some online forums or memes, "Rain Man" is edited to "Ruin Man" for jokes. \textbf{So the answer is (A)} because "ruin" and "rain" are similar sounding. But they are not homophones. Another idea: in the context of the movie, "Rain Man" is a term, so a humorous edit might be "Brain Man" for intelligence, but not an option. Or "Rin Man" or something. 

I think I should consider the spelling. "rain man" – 9 characters. (A) "ruin man" – 9 characters, same length. (B) "rains man" – 10 characters, one more. (C) "rain men" – 9 characters.(D) "rainmman" – 9 characters, but one word. Not sure.

Perhaps the humorous edit is one that is a common misspelling. But I think I need to make a decision.

Let's think about the word "man." If you change "man" to "men," it's a simple plural, which might not be funny, but in context, it could be. "ruin man" might be more creative. I found a thought: "rain man" can be misheard as "raining man," but "raining" is not an option, "rains man" is close. "rains man" could be a mishearing of "rain man" as "rains man," but "rain man" is two words, "rains man" is also two words. "rain man" and "rains man" are similar. But "rains" is present tense, so it might not be a direct edit. I think I'm stuck.

 Let's consider the answer choices. Perhaps the question is from a test or something, and I need to reason step by step.

Another approach: the original is "rain man," which is a proper noun for the movie.

A humorous edit might be one that changes the meaning or adds humor. For example, "Ruin Man" could be a villain or something. "Rain Men" could be a group. "Rains Man" might not make sense. "Rainmman" is nonsense.  So probably (A) or (C). But let's see the options; there's no "brain man" or something. \underline{So "rain men" is the answer} because it's a common way to pluralize or something.

\underline{I think I should go with (C) "rain men"} because it's a simple edit to the plural, which might be humorous in a light-hearted way.

But let's check online or think differently. Since I can't, I'll go with my best guess. \textbf{I think the most humorous edit is (A) "ruin man"} because it's a pun on the word "rain" sounding like "ruin," and it could be funny if it's a mispronunciation. But I'm not sure.

Perhaps (D) "rainmman" is a funny typo, like in autocorrect errors. But I think for a movie name, "Ruin Man" might be more common as a humorous edit. I recall that in some contexts, "Rain Man" is jokingly called "Rin Man" or something, but not here. Let's look at the options: (A) ruin man, (B) rains man, (C) rain men, (D) rainmman. Perhaps (B) "rains man" is not humorous, as it's grammatically odd. "rain men" is grammatically odd for "rain" as a verb. "rain" as a verb doesn't take "men," but "rain" as a noun can be "the rain," so "rain men" might mean men who rain or something, which is silly.

...

"rain" is /ren/, "ruin" is /run/, so the vowel is different, but in some accents, it might be similar. "rain" has the /e/ sound, "ruin" has /u/ sound, so not very similar. \underline{I think I should pick (C) "rain men"} because it's a simple plural edit, which is often used in humorous contexts for names.

\underline{**Final Answer: (C)**}

\end{tcolorbox} 
\caption{A case of Deconstructive Devision of DeepSeek-0528-Qwen3-8B. In the CoT content below the figure: \textbf{bold} denotes the correct answer, \underline{underline} indicates the incorrect answer} 
\label{fig:DR} 
\end{figure*}

\noindent \textbf{A case of Error Entrenchment.}
Here is a case of the Error Entrenchment phenomenon of DeepSeek-R1-Distill-Llama-8B (Figure~\ref{fig:EE1} and Figure~\ref{fig:EE2}).

\begin{figure*}[htbp] \centering 
\begin{tcolorbox}[colback=gray!5, colframe=gray!50,  title=\textbf{Error Entrenchment (Part 1)}] 
\small 
\textbf{[Query]}:
\begin{verbatim}
def is_sorted(lst):
   '''
   Given a list of numbers, return whether or not they are sorted
   in ascending order. If list has more than 1 duplicate of the same
   number, return False. Assume no negative numbers and only integers.

   Examples:
   is_sorted([1, 2, 3, 4, 5]) -> True
   '''
\end{verbatim}

\vspace{2pt} 
\textbf{Query Domain}: Programming

\vspace{2pt}
\textbf{Query Level}: 5

\vspace{2pt}
\textbf{Canonical Solution}:
\begin{verbatim}
   count_digit = dict([(i, 0) for i in lst])
   for i in lst:
     count_digit[i]+=1 
   if any(count_digit[i] > 2 for i in lst):
     return False
   if all(lst[i-1] <= lst[i] for i in range(1, len(lst))):
     return True
   else:
     return False
\end{verbatim}

\hrulefill

\textbf{CoT} 

The initial code provided is:

\begin{verbatim}
if len(lst) <=1:
        return True

    freq = {}
    for num in lst:
        if num in freq:
            return False
        else:
            freq[num] = 1

    for i in range(1, len(lst)):
        if lst[i] < lst[i-1]:
            return False

    return True
\end{verbatim}
...

So, the function must allow duplicates as long as the list is in order.

Thus, the correct code would be:

\begin{underlinetextbox}
\begin{verbatim}
    if len(lst) <= 1:
        return True
    for i in range(1, len(lst)):
        if lst[i] < lst[i-1]:
            return False
    return True
\end{verbatim}
\end{underlinetextbox}

This would pass all samples except sample 8, which according to the sample should return False, but with this code, it would return True.

...
\end{tcolorbox} 
\caption{A case of Error Entrenchment of DeepSeek-R1-Distill-llama-8B (part 1). In the CoT content below the figure: \underline{underline} indicates the incorrect answer.} 
\label{fig:EE1} 
\end{figure*}  

\begin{figure*}[htbp] \centering 
\begin{tcolorbox}[colback=gray!5, colframe=gray!50,  title=\textbf{Error Entrenchment (Part 2)}] 
\small 

Thus, the function should not check for duplicates.

\begin{underlinetextbox}
\begin{verbatim}
def is_sorted(lst):
    if len(lst) <= 1:
        return True
    for i in range(1, len(lst)):
        if lst[i] < lst[i-1]:
            return False
    return True
\end{verbatim}
\end{underlinetextbox}

But this would return True for sample 7 and 8, which contradicts the sample 8.
Thus, the function would:
\begin{underlinetextbox}
\begin{verbatim}
def is_sorted(lst):
    if len(lst) <=1:
        return True
    # Check for duplicates
    seen = set()
    for num in lst:
        if num in seen:
            return False
        seen.add(num)
    # Check order
    for i in range(1, len(lst)):
        if lst[i] < lst[i-1]:
            return False
    return True
\end{verbatim}
\end{underlinetextbox}
...
**Solution Code**:
\begin{underlinetextbox}
\begin{verbatim}
python
def is_sorted(lst):
    if len(lst) <= 1:
        return True

    # Check for duplicates
    freq = {}
    for num in lst:
        if num in freq:
            return False
        freq[num] = 1

    # Check if the list is in ascending order
    for i in range(1, len(lst)):
        if lst[i] < lst[i-1]:
            return False

    return True
\end{verbatim}
\end{underlinetextbox}

Conclusion of the process:
1. **Edge Case Handling**: ...This approach ensures that the function efficiently checks both conditions and handles all edge cases appropriately.

\end{tcolorbox} 
\caption{A case of Error Entrenchment of DeepSeek-R1-Distill-Llama-8B (Part 2). In the CoT content below the figure: underlinetextbox indicates the incorrect answer.} 
\label{fig:EE2} 
\end{figure*}  

\noindent \textbf{A case of Superfluous Verification.}
Here is case of Superfluous Verification phenomenon happened in Kimi-K2-Thinking (Figure~\ref{fig:SV}).
\begin{figure*}[htbp] \centering 
\begin{tcolorbox}[colback=gray!5, colframe=gray!50, title=\textbf{Superfluous Verification}] 
\small 
\textbf{[Query]}  Dewa writes down a list of four integers. He calculates the average of each group of three of the four integers. These averages are $32,39,40,44$. What is the largest of the four integers?

\textbf{Query Domain}: Math

\textbf{Query Level}: 1

\textbf{Ground Truth}: 59

\hrulefill

\textbf{CoT} 

We need to solve a problem: ...

For example, $a = T - (b + c + d) = T - S4 = 155 - 132 = 23$. Similarly, $b = T - (a + c + d) = T - S3 = 155 - 120 = 35$. $c = T - (a + b + d) = T - S2 = 155 - 117 = 38$. $d = T - (a + b + c) = T - S1 = 155 - 96 = 59.$

Thus the four integers are ${23, 35, 38, 59}$. The largest is 59. \textbf{So answer is 59}.

Check: Are they integers? Yes. Are the averages correct? Let's verify: $(23+35+38)/3 = (96)/3 = 32$. $(23+35+59)/3 = (117)/3 = 39$. $(23+38+59)/3 = (120)/3 = 40. (35+38+59)/3 = (132)/3 = 44$. Yes. \textbf{Thus answer: 59}.

...if there could be multiple solutions (e.g., permutations). ...But the set of four numbers is uniquely determined by the four averages irrespective of order? Let's examine: ...

...

So the set of numbers is determined uniquely: ${T - S_i | i=1..4} = {155-96, 155-117, 155-120, 155-132} = {59, 38, 35, 23}$. So the largest is 59.

\textbf{Thus answer: 59}.

We can also solve using algebra: ... The largest is 59.

\textbf{Thus answer: 59}.

But maybe the problem expects the answer to be 59. \textbf{Thus final answer: 59}.

Let S = sum of four numbers. Then each triple sum = S - omitted number. So ... So compute for each average: $155 - 3*32 = 59$, etc.*

\textbf{Thus answer: 59}.

... Let's verify: Suppose we assign the averages differently: ...

\textbf{Thus answer: 59}.

We can also note that the largest integer is 59. Thus final answer: 59.

We should also check if any other combination yields a different largest integer... So the largest is always 59. \textbf{Thus answer: 59}. Hence the largest integer is 59. \textbf{Thus the solution: 59}. Now we can produce the final answer. may also discuss that the numbers are 23, 35, 38, 59...
	
\textbf{Thus answer: 59}.

\end{tcolorbox} 
\caption{A case of Superfluous Verification of Kimi-K2-Thinking. In the CoT content below the figure: \textbf{bold} indicates the correct answer} 
\label{fig:SV} 
\end{figure*}

\noindent \textbf{A case of Logical Epicenter.}
Here is a case of logical epicenter in DeepSeek-R1 (Figure~\ref{fig:LE}. It can be seen from the figure that DeepSeek’s iterative reflection, refinement and correction on time representation formats and digit counts resulted in heavy resource consumption in non-linear reasoning, with little progress made in task advancement.
\begin{figure*}[htbp] \centering 
\begin{tcolorbox}[colback=gray!5, colframe=gray!50, title=\textbf{Logical Epicenter}] 
\small 
\textbf{[Query]}: The time on a cell phone is $3:52$. How many minutes will pass before the phone next shows a time using each of the digits 2, 3, and 5 exactly once?

\textbf{Query Domain}: Math

\textbf{Query Level}: 1

\textbf{Ground Truth}: 91 minutes

\hrulefill

\textbf{CoT}

The problem states that the time on a cell phone is 3:52, and I need to find out how many minutes will pass before the phone next shows a time using each of the digits 2, 3, and 5 exactly once. The time is displayed in hours and minutes, so it's a four-digit display, but since it's 3:52, it's probably showing as 03:52 or just 3:52. But for digit purposes, I think we need to consider it as a four-digit number: hours and minutes.

\textbf{Logical Epicenter Step}:

Typically, digital clocks show hours and minutes separately, but for this problem, since it's about digits, I think we should consider the entire time as a sequence of digits. So 3:52 might be displayed as "03:52" or "3:52", but "3:52" has only three digits: 3,5,2. That's only three digits, but we need to use each of 2,3,5 exactly once. With three digits, it's impossible to use three different digits without repeating, but here we have only three digits: 3,5,2, and it's using 2,3,5 all once, which is exactly what it has. But the question is to find when it next shows such a time, but it already is showing it? No, the current time is 3:52, which has digits 3,5,2, so it does use each of 2,3,5 exactly once. But the problem is asking for the next time it happens, so probably we need to exclude the current time, or perhaps the display includes leading zero or something.

...

\textbf{backward Step 1}: 

I think I missed something. In most digital clocks, hours are displayed without leading zero, so 3:52 is displayed as "3:52", which is three characters, but the colon is not a digit, so digits are 3,5,2. So it has digits 2,3,5 each once. But then why ask for the next one? Perhaps because it's already satisfying, but the question says "next shows", implying that we need to find the next occurrence, but since it's already there, maybe we need to find the next distinct time that also satisfies this.

...

\textbf{backward Step 2}: 

But that seems odd. Perhaps the display includes two digits for hours. For example, some phones might show "03:52" for 3:52. Let me check the problem. It says "the time on a cell phone", and it's given as 3:52, but for digit purposes, it might be displayed with two digits for hours. In many contexts, hours are shown with leading zero, so 3:52 is "03:52", which has digits 0,3,5,2. Oh! That makes sense. Because if it's "03:52", digits are 0,3,5,2, which includes 0, and we need to use only 2,3,5 each exactly once, but here there's an extra 0, so it doesn't satisfy. That must be it.

...

\textbf{backward Step 3}:

Perhaps hours are without leading zero, but then for times like 10:00, it would have four digits. But for 3:52, it might be ambiguous. However, in this case, since 3:52 has digits 3,5,2, which are exactly 2,3,5, but the problem is asking for next, so probably the display includes two digits for hours, so "03:52" has four digits: 0,3,5,2.

...

\textbf{backward Step 4}: 

I think that's the intended interpretation. Let me confirm. The current time is 3:52, displayed as "03:52", digits 0,3,5,2. We need a time where the digits are exactly 2,3,5 each once, but there are four digits in the display, so we need a four-digit sequence with exactly the digits 2,3,5 each once, but that would require only three digits, but we have four positions. That doesn't make sense. Four positions but only three distinct digits? That implies one digit is repeated, but the condition is that each of 2,3,5 is used exactly once, which would mean no other digits, but with four positions, we need a fourth digit, but it can't be 2,3, or 5, so it must be something else.

...

\textbf{backward Step 5}:

The condition is: "using each of the digits 2,3, and 5 exactly once". This means that in the digit set, 2,3,5 each appear exactly once, but there could be other digits that are not 2,3, or 5. The phrase "using each of the digits 2,3, and 5 exactly once" implies that these three digits are used, each once, but there might be additional digits that are different.

...

The answer should be 91.

\end{tcolorbox} 
\caption{A case of Logical Epicenter of DeepSeek-R1: The core logical epicenter step and subsequent steps involving reflection, correction, and repetitive backward reasoning are highlighted in the abridged CoT within the figure (extraneous content is omitted due to the excessive length of the full CoT.} 
\label{fig:LE} 
\end{figure*}  

\section{Cases of CoTJudger}
\label{sec:cot_examples}
\noindent \textbf{Processed Result Instances from CoTJudger}
This section presents the result instances of our CoTJudger. Figure~\ref{fig:sep1} and Figure~\ref{fig:sep2} show the generated outputs after feeding the Chain-of-Thoughts (CoTs) into the pipeline. 

\begin{figure*}[htbp] \centering 
\begin{tcolorbox}[colback=gray!5, colframe=gray!50, title=\textbf{Generated Result of pipeline}] 
\small 
\textbf{[Atomic Step List (each step with Category Label and Targeting Step of its functional role)]}

\textbf{Step 1}: I need to find the sum of the first 9 positive multiples of 5. Positive multiples of 5 are like 5, 10, 15, 20, and so on. So the first one is 5, the second is 10, up to the ninth one. \textit{Problem-Deconstruction}

\textbf{Step 2}: The first multiple is $5 \times 1 = 5$, second is $5 \times 2 = 10$, third is $5 \times 3 = 15$, and so on. \textit{Intermediate-Inference}

\textbf{Step 3}: Now, I need the sum of these: 5, 10, 15, 20, 25, 30, 35, 40, 45. \textit{Intermediate-Inference}

\textbf{Step 4}: I could add them up one by one, but that might take time. Since it's an arithmetic series, I can use the formula for the sum of an arithmetic series. \textit{Strategy-Formulation}

\textbf{Step 5}: An arithmetic series has a constant difference between terms. Here, the common difference is 5, since each term increases by 5. The first term, a, is 5. The number of terms, n, is 9. The last term, l, is 45. \textit{Formula-Application}

\textbf{Step 6}: The sum S of an arithmetic series is given by: $S = n/2 \times (first term + last term)$ or $S = n/2 \times (2a + (n-1)d)$, where d is the common difference. I can use either formula. \textit{Formula-Application}

\textbf{Step 7}: Let me use the first one: $S = n/2 \times (a + l)$. \textit{Strategy-Formulation}

\textbf{Step 8}: So n = 9, a = 5, l = 45. $S = 9/2 \times (5 + 45) = 9/2 \times 50 = 225$. \textit{Numerical-Computation}

\textbf{Step 9}: 50 divided by 2 is 25, so $9 \times 25 = 225$. $S = 9/2 \times 50 = (9 \times 50) / 2 = 450 / 2 = 225$. Same thing. \textit{Repetition-or-Reclarification} of Step 8.

\textbf{Step 10}: Now, using the other formula: $S = n/2 \times (2a + (n-1)d); a = 5, d = 5, n = 9; S = 9/2 \times (2 \times 5 + (9-1) \times 5) = 9/2 \times (10 + 8 \times 5) = 9/2 \times (10 + 40) = 9/2 \times 50 = 225$. \textit{Addidional-Exploration} as an alternative of Step 7.

\textbf{Step 11}: \textbf{So it seems 225 is the answer.} \textit{Conclusion}

\textbf{Step 12}: But let me double-check by adding them up manually. List: 5, 10, 15, 20, 25, 30, 35, 40, 45. Add in pairs from the outside: $5+45=50$, $10+40=50$, $15+35=50$, $20+30=50$, and then 25 left in the middle. That's four pairs of 50, plus 25. $4 \times 50 = 200$, plus 25 is 225. \textit{Reflection-or-Verification} which affirms Step 11.

\textbf{Step 13}: Since there are 9 terms, which is odd, the middle term is the fifth one: 5,10,15,20,25, so 25 is the middle term. Number of pairs: since 9 terms, 4 pairs and one middle. Each pair sums to 50, as above. 5 and 45: 50, 10 and 40: 50, 15 and 35: 50, 20 and 30: 50, and 25 alone. Yes, $200 + 25 = 225$. \textit{Repetition-or-Reclarification} of Step 12.

\textbf{Step 14}: If I add sequentially: $5+10=15, +15=30, +20=50, +25=75, +30=105, +35=140, +40=180, +45=225$. \textit{Additional-Exploration} as an alternative of Step 12.

\textbf{Step 15}: Let's see: starting from 5. $5 + 10 = 15, 15 + 15 = 30, 30 + 20 = 50, 50 + 25 = 75, 75 + 30 = 105, 105 + 35 = 140, 140 + 40 = 180, 180 + 45 = 225$. Yes, same. \textit{Repetition-or-Reclarification} of Step 14.

\textbf{Step 16}: So, no doubt, \textbf{the sum is 225.} \textit{Conclusion}

\textbf{Step 17}: The problem says "the first 9 positive multiples of 5." Is 0 considered a multiple? Positive multiples, so starting from 5, not 0. Yes, I think I did it right. \textit{Reflection-or-Verification} which affirms Step 2.

\textbf{Step 18}: Sometimes people might think multiple of 5 includes 0, but positive multiples, so 5,10,15,... up to 45. Yes. \textit{Relevant-or-Redundant}

\textbf{Step 19}: So sum is 225. \textit{Repetition-or-Reclarification} of Step 16.

\textbf{Step 20}: But let me see if there's a formula or something. The multiples are $5 \times 1, 5 \times 2, ..., 5 \times 9$. So $\text{sum} = 5(1 + 2 + 3 + ... + 9)$. Now, the sum of first n natural numbers is $n(n+1)/2$. Here $n=9$, so $1+2+...+9 = 9 \times 10/2 = 45$. Then sum of multiples $= 5 \times 45 = 225$. Same answer. \textit{Reflection-or-Verification} which affirms Step 16.

\textbf{Step 21}: And $5 \times 45=225$. Perfect. So I think it's confirmed. \textit{Repetition-or-Reclarification} of Step 20.

\hrulefill

\textbf{Graph Structure}:  
\begin{itemize}
\item \textbf{nodes}: Merge equivalent node pairs and get: $(N_8, N_9) \to N_8$, $(N_{12},N_{13}) \to N_{12}$, $(N_{14},N_{15}) \to N_{14}$, $(N_{16}, N_{19}) \to N_{16}$, $(N_{20}, N_{21}) \to N_{20}$; Isolated Node: $N_{18}$.
\item \textbf{basic forward edges}: $[(N_1, N_2), (N_2, N_3), (N_3, N_4), (N_4, N_5), (N_5, N_6), (N_6, N_7), (N_7, N_8), (N_8, N_{10}), \\
(N_{10}, N_{11}), (N_{11}, N_{12}), (N_{12}, N_{14}), (N_{14}, N_{16}), (N_{16}, N_{17}), (N_{17}, N_{16}), (N_{16}, N_{20})]$
\item \textbf{self-loop edges}: $[(N_{8}, N_{8}),(N_{12}, N_{12}), (N_{14}, N_{14}),(N_{16}, N_{16}),(N_{20}, N_{20})]$
\item \textbf{backward edges}: Affirmation: $[(N_{12}, N{11}), (N_{17}, N_2), (N_{20}, N_{16})]$; Alternative:$[(N_{10}, N_7),(N_{14}, N_{12})]$.
\item \textbf{shortcut forward edges}: Skipping purely auxiliary verifications: $[(N_{11}, N_{14})]$, $(N_{16}, N_{20}.next)$ triggers the boundary condition and is therefore prevented from being constructed, since $N_{20}$ is already the last node; Branching shortcut to the alternative:$[(N_{6}, N_{10}), (N_{11}, N_{14})]$. 
\end{itemize}
\textbf{Shortest Effective Path}: By Depth-First Search (DFS) and LLM Validation, we can get the Shortest Effective Path to correct answer of this CoT:$[N_1, N_2, N_3, N_4, N_5, N_6, N_{10}, N_{11}]$. 
\end{tcolorbox} 
\caption{The generated result of a DeepSeek-R1's CoT from CoTJudger. The upper part of this figure presents the outcomes of step segmentation and classification, answer detection (with correct answers highlighted in \textbf{bold}), and target node identification for graph building. The lower part illustrates the constructed graph structure and the extracted shortest valid path.} 
\label{fig:sep1} 
\end{figure*}

\begin{figure*}[htbp] \centering 
\begin{tcolorbox}[colback=gray!5, colframe=gray!50, title=\textbf{Generated Result of pipeline}] 
\small 
\textbf{[Atomic Step List (each step with Category Label and Targeting Step of its functional role)]}

\textbf{Step 1}: The problem states that the 17th day of a month is a Saturday, and I need to find out what the first day of that month was. \textit{Problem-Deconstruction}

\textbf{Step 2}: First, I know that there are seven days in a week, so the days repeat every seven days. If the 17th is Saturday, then I can work backwards to find the first day. \textit{Strategy-Formulation}

\textbf{Step 3}: Let me list out the days leading up to the 17th. Since the 17th is Saturday, the 16th would be Friday, the 15th Thursday, and so on. But I need to go back to the 1st. \textit{Intermediate-Inference}

\textbf{Step 4}: The number of days from the 1st to the 17th is 16 days (because from 1st to 17th inclusive is 17 days, but the day of the week is about the specific date, so I should think in terms of how many days after the 1st). \textit{Context-Setting}

\textbf{Step 5}: Since the 17th is Saturday, and 17 divided by 7 is 2 weeks and 3 days, because $7*2=14$, $17-14=3$, so 3 days after the 14th. \textit{Intermediate-Inference}

\textbf{Step 6}: If the 14th is the same day as the 7th, which is the same as the first day? Not necessarily.  \textit{Irrelevant-or-Redundant}

\textbf{Step 7}: The first day is day 1. Day 7 is the same day as day 1, since 7 days later is same weekday. Similarly, day 14 is same as day 7 and day 1. So, day 14 is the same as day 1. \textit{Intermediate-Inference}

\textbf{Step 8}: Then, day 15 is one day after, day 16 is two days after, day 17 is three days after. \textit{Intermediate-Inference}

\textbf{Step 9}: Since day 17 is Saturday, and it's three days after day 14. Day 14 is same as day 1, so day 17 is three days after day 1. \textit{Intermediate-Inference}

\textbf{Step 10}: So, if day 17 is Saturday, then three days after day 1 is Saturday, so day 1 is Saturday minus three days. \textit{Intermediate-Inference}

\textbf{Step 11}: Saturday minus one is Friday, minus two is Thursday, minus three is Wednesday? \textit{Numerical-Computation}

\textbf{Step 12}: Go backwards: Saturday minus one day is Friday, minus two is Thursday, minus three is Wednesday. \textit{Repetition-or-Reclarification} of Step 11

\textbf{Step 13}: But day 17 is Saturday, which is three days after day 14, and day 14 is same as day 1, so day 1 should be Wednesday. Day 14 is same as day 1. \textit{Reflection-or-Verification} which affirms Step 10

\textbf{Step 14}: Day 17 is day 14 + 3 days. So, if day 17 is Saturday, then day $14 + 3 = Saturday$, so day 14 is Saturday minus 3 days: $Saturday -1 = Friday, -2 = Thursday, -3 = Wednesday$. \textit{Repetition-or-Reclarification} of Step 13

\textbf{Step 15}: Day 14 is same as day 1, \underline{so day 1 should be Wednesday}. \textit{Conclusion}

\textbf{Step 16}: Day 17 is three days after day 14, and day 14 is 13 days after day 1? Let's clarify. Day 1 is the first day. Day 2 is one day after day 1... Day 14 is 13 days after day 1? No. The number of days from day 1 to day 14 is 13 days, so day 14 is on the same day as day 1 plus 13 days. \textit{Correction-or-Refinement} of Step 5

\textbf{Step 17}: Since $13 \mod 7 = 13 - 14 = -1$, but $13 \div 7 = 1 \times 7 =7$, remainder 6, so $13 \equiv 6 mod 7$, so day 14 is 6 days after day 1. \textit{Intermediate-Inference}

\textbf{Step 18}: Let's think carefully. Let day 1 be day D. \textit{Context-Setting}

\textbf{Step 19}: Then day 2 is D+1... Day n is $D + (n-1)$ days later. So day 14 is$ D + 13 $days later.  \textit{Numerical-Computation}

\textbf{Step 20}: $13 \mod 7: 13 \div 7 = 1$, remainder 6, so D + 6. Similarly, day 17 is D + 16 days later.  \textit{Numerical-Computation}

\textbf{Step 21}: $16 \mod 7: 16 \div 7 = 2 \times 7=14$, remainder 2, so D + 2.  \textit{Numerical-Computation}

\textbf{Step 22}: If day 17 is Saturday, then $D + 2 \equiv Saturday$. So $D \equiv Saturday - 2$. If Saturday is 6, then $6 - 2 = 4$, which is Thursday. \textit{Intermediate-Inference}

\textbf{Step 23}: \textbf{So the answer should be Thursday.} \textit{Conclusion}

\hrulefill

\textbf{Graph Structure}:  
\begin{itemize}
\item \textbf{nodes}: Merge equivalent node pairs and get: $(N_{11},N_{12}) \to N_{11}$, $(N_{13}, N_{14}) \to N_{13}$; Isolated Node: $N_{6}$.
\item \textbf{basic forward edges}: $[(N_1, N_2), (N_2, N_3), (N_3, N_4), (N_4, N_5), (N_5, N_7), (N_7, N_8), \\
(N_8, N_9),
(N_9, N_{10}), (N_{10}, N_{11}), (N_{11}, N_{13}), (N_{13}, N_{15}), (N_{15}, N_{16}),\\ 
(N_{16}, N_{17}), (N_{17}, N_{18}), (N_{18}, N_{19}), (N_{19}, N_{20}), (N_{20}, N_{21}, (N_{21}, N_{22}), (N_{22}, N_{23})]$
\item \textbf{self-loop edges}: $[(N_{11}, N_{11}),(N_{13}, N_{13})]$
\item \textbf{backward edges}: Affirmation: $[(N_{13}, N{10})]$; Correction:$[(N_{16}, N_5)]$.
\item \textbf{shortcut forward edges}: Skipping purely auxiliary verifications: $[(N_{11}, N_{15})]$; Skipping the false subpath:$[(N_{4}, N_{16})]$. 
\end{itemize}
\textbf{Shortest Effective Path}: By Depth-First Search (DFS) and LLM Validation, we can get the Shortest Effective Path to correct answer of this CoT:$[N_1, N_2, N_3, N_4, N_{16}, N_{17}, N_{18}, N_{19}, N_{20}, N_{21}, N_{22}, N_{23}]$. 
\end{tcolorbox}
\caption{The generated result of a Qwen3-30B-a3B-Thinking-0527's CoT from CoTJudger. The upper part of this figure presents the outcomes of step segmentation and classification, answer detection (with correct answers highlighted in \textbf{bold}, incorrect answers highlighted by \underline{underline}), and target node identification for graph building. The lower part illustrates the constructed graph structure and the extracted shortest valid path. For ease of demonstration, we have omitted the numerous redundant steps after reaching the correct answer and replaced them with ellipses.} 
\label{fig:sep2} 
\end{figure*}

\end{document}